\documentclass[journal]{IEEEtran}
\usepackage{times}
\usepackage{epsfig}
\usepackage{graphicx}
\usepackage{amsmath}
\usepackage{amssymb}
\usepackage{mathtools}
\usepackage{algorithm}
\usepackage{algorithmic}
\usepackage{graphics}
\usepackage{subfigure}
\usepackage{multirow}
\usepackage{booktabs}
\usepackage{url}
\usepackage{amsthm}
\usepackage{wrapfig}
\usepackage{xcolor}

\usepackage{makecell}
\usepackage{ragged2e}

\usepackage{xspace}
\newcommand*{\eg}{\emph{e.g.}\@\xspace}
\newcommand*{\ie}{\emph{i.e.}\@\xspace}
\newcommand*{\etc}{\etc{etc.}\@\xspace}

\DeclareRobustCommand
      \Compactcdots{\mathinner{\cdotp\mkern-2mu\cdotp\mkern-2mu\cdotp}}

\usepackage{cite}

\begin{document}
%
\title{A Closer Look at Branch Classiﬁers of Multi-exit Architectures}
%
%
%

\author{Shaohui Lin$^{*}$,~\IEEEmembership{Member,~IEEE,}
        Bo Ji,
        Rongrong Ji,~\IEEEmembership{Senior Member,~IEEE,}
        and Angela Yao
\thanks{S. Lin (Corresponding Author) is with the School of Computer Science and Technology, East China Normal University, 200062, China. (Email: shlin@cs.ecnu.edu.cn)}
\thanks{B. Ji is with the School of Computing, National University of Singapore, 117417, Singapore.}
\thanks{R. Ji is with the Lab of Media Analytics and Computing, School of Informatics, Xiamen University, 361005, China.}
\thanks{A. Yao is with the School of Computing, National University of Singapore, 117417, Singapore.}
}

%
%

\markboth{Submission to Computer Vision and Image Understanding}%
{Shell \MakeLowercase{\textit{et al.}}: Bare Demo of IEEEtran.cls for IEEE Journals}

\maketitle

\begin{abstract}
Multi-exit architectures consist of a backbone and branch classifiers that offer shortened inference pathways to reduce the run-time of deep neural networks.  
In this paper, we analyze different branching patterns that vary in their allocation of computational complexity for the branch classifiers.  Constant-complexity branching keeps all branches the same, while complexity-increasing and complexity-decreasing branching place more complex branches later or earlier in the backbone respectively.  Through extensive experimentation on multiple backbones and datasets, we find that complexity-decreasing branches are more effective than constant-complexity or complexity-increasing branches, which achieve the best 
accuracy-cost trade-off.  We investigate a cause by using knowledge consistency to probe the effect of adding branches onto a backbone.  Our findings show that complexity-decreasing branching yields the least disruption to the feature abstraction hierarchy of the backbone, which explains the effectiveness of the branching patterns.
\end{abstract}

\begin{IEEEkeywords}
multi-exit architectures, knowledge consistency, branch classifiers, model compression and acceleration
\end{IEEEkeywords}

%
\IEEEpeerreviewmaketitle

%
%
%
%
\section{Introduction}
\label{intro}
\IEEEPARstart{C}{onvolutional} Convolutional neural networks (CNNs) for visual recognition have become progressively more accurate, yet also bigger, deeper, and slower. State-of-the-art CNNs 
such as ResNet-152 \cite{he2016deep} require tens of billions FLOPs to classify a single image, which brings serious limitations for deploying models on resource-constrained platforms. 
Efficient network design \cite{sandler2018mobilenetv2,han2020ghostnet} and acceleration methods such as pruning~\cite{han2015learning,he2019asymptotic,wang2021soft}, quantization \cite{jacob2018quantization,zhang2019memristive} and low-rank decomposition~\cite{lin2018holistic,ruan2020edp,denton2014exploiting} can improve CNN inference efficiency.  However, it is inevitable that increasing efficiency leads to a decrease in accuracy.  
Multi-exit architectures  \cite{huang2017multi,scardapane2020should,passalis2020efficient,bicici2021conditional} 
try to minimize this trade-off by allocating computational budget dynamically at inference time.  From a backbone network, branch classifiers are added at intermediate layers to 
exit samples according to classification difficulty.  Such a paradigm assumes that ``easy'' samples can be classified successfully by the branches without passing through the entire network. If these samples can be found reliably, then inference budget can be allocated accordingly.

In developing multi-exit CNNs, feature reuse~\cite{huang2017multi,passalis2020efficient}, training strategy~\cite{li2019improved,phuong2019distillation,zhang2019scan} and termination policies \cite{bolukbasi2017adaptive,teja2018hydranets,shen2020fractional} of branch classifiers have been studied.  Various branch architectures have been proposed, ranging from simple 
convolutions~\cite{huang2017multi,yang2020resolution,Yang2020mutualnet} 
to highly elaborate structures~\cite{zhang2019scan}.  
Yet as we will show, branches 
exhibit complex interactions and interference effects with each other and the backbone. When done naively, this is detrimental and may deteriorate network performance downstream from the branch.  

\begin{figure*}[t]
\centering
    \includegraphics[scale = 0.31]{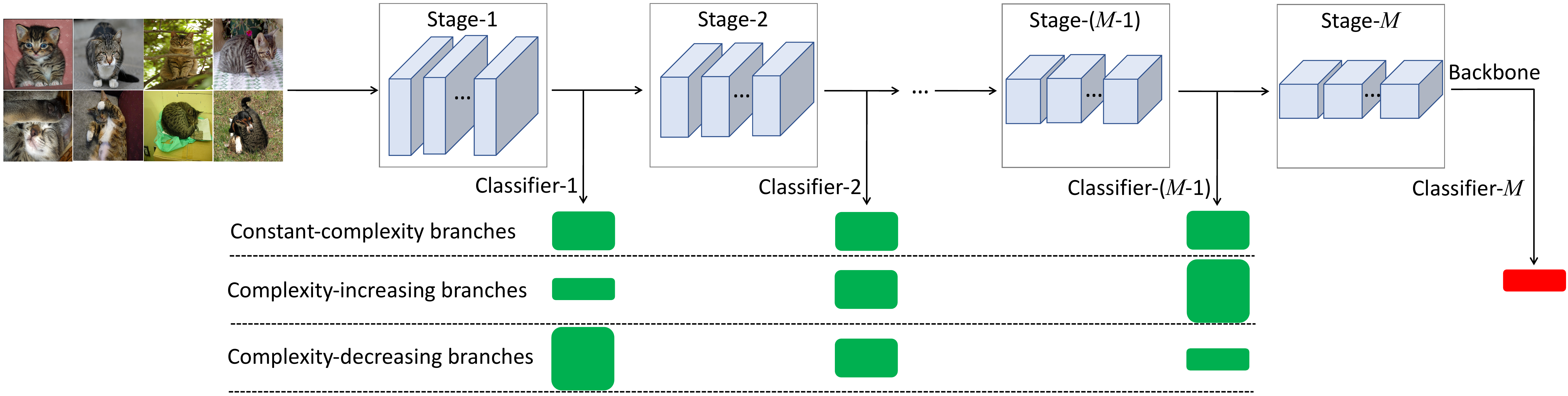}
\caption{Proposed multi-exit framework with varying branching patterns. Green rectangles represent branch classifiers; the longer the rectangle, the more complex the branch classifier. Constant-complexity branches have the same classifier on all branches, while complexity-increasing and complexity-decreasing branches have classifiers which get gradually deeper or shallower respectively. In this work, we fix the total FLOP budget over all branches and explore the performance of different branching patterns. Experiments show that complexity-decreasing branches achieve the best speed-accuracy trade-off.
}
\label{fig:framework}
~\vspace{-1em}
\end{figure*}

In this paper, we investigate the role of branch complexity for multi-exit frameworks under fixed total FLOP budgets and look at three branching patterns: constant-complexity, complexity-increasing and complexity-decreasing (see Fig.~\ref{fig:framework}).  Through extensive experimentation on multiple pre-trained backbones, we find that~\emph{complexity-decreasing} branches offers the best cost-accuracy trade-off. To investigate the cause, we introduce the use of knowledge consistency~\cite{liang2019knowledge} as a tool to diagnose the impact of the different branching patterns on a backbone network.  
Our findings show that branching earlier results in architectures with more consistent knowledge to backbone baselines.  They are likely to have less disruption to the feature abstraction hierarchy.  As such, we advocate using a complexity-decreasing branching pattern for practitioners developing multi-exit frameworks. 
This finding is consistent to the suggestion that later branch points should have fewer layers in the BranchyNet~\cite{teerapittayanon2016branchynet}, while our complexity-decreasing branching classifiers are elaborately designed, analyzed and evaluated comprehensively under the setting of a fixed total FLOPs budget. Moreover, our complexity-decreasing branching classifiers are architecture-agnostic to effectively construct early-exit framework for various network architectures (e.g. DenseNets~\cite{huang2017densely} and MobileNetV2 \cite{sandler2018mobilenetv2})), rather than the simpler networks like LeNet~\cite{lecun1998gradient} and AlexNet~\cite{krizhevsky2012imagenet}.

During training, we find that knowledge distillation can further improve branch classifier accuracy.  We formulate a new online weighted ensemble as the teacher which achieves state-of-the-art performance on various backbones and datasets. 
Compared to other multi-exit classifiers such as MSDNet~\cite{huang2017multi} and SCAN~\cite{zhang2019scan}.  On the same ResNet backbones, we improve upon their adaptive accuracy by as much as 3.3\% with 0.3B less FLOPS.

Our contributions can be summarized as:
\begin{itemize}
\item[1.] We propose a novel multi-exit framework to investigate different branching patterns and find that a complexity-decreasing pattern 
achieves the best accuracy-speed trade-off under the same total FLOP budget of branches.
\item[2.] We introduce a method to compare branching patterns via knowledge consistency~\cite{liang2019knowledge}. 
Results suggest that complexity-decreasing branches have less disruption to the feature abstraction hierarchy in the backbone as an explanation for its effectiveness. 
\item[3.] 
Using a weighted ensemble for distillation training with complexity-decreasing branches, we achieve state-of-the-art results on ImageNet with higher adaptive accuracies and lower average FLOPS 
when compared to previous methods~\cite{huang2017multi,zhang2019scan}. 
\end{itemize}

\section{Related work}\label{sec:rw}

\textbf{Early exit.}
The seminal BranchyNet~\cite{teerapittayanon2016branchynet} was first to add early-exit classifiers to a deep network.  
Bolukbasi~\emph{et al.}~\cite{bolukbasi2017adaptive} extended this concept to multiple networks and 
proposed a 
network selection policy for cascades of pre-trained models. 
As the cascade did not reuse features across classifiers, it required substantial storage overhead. 
Other improvements on the early-exit framework include feature reuse~\cite{huang2017multi,wang2020glance}, distillation-based training~\cite{phuong2019distillation,li2019improved,zhang2019scan} and data augmentation~\cite{hu2020triple}. 
However, branch classifier design was often overlooked; simple convolutions were used and branches were kept the same~\cite{huang2017multi,li2019improved}.  Recently,~\cite{zhang2019scan} introduced varying branches with highly elaborate branch structures, though their focus was to construct scalable networks. More related works for early exiting can be further reviewed in~\cite{han2021dynamic}.
We are inspired by them to go beyond simple convolutions for branch architectures and systematically analyze different branching patterns under fixed total FLOP budgets. 

\textbf{Dynamic layer and channel selection.}
Recently, networks whose parts are selectively executed or skipped at test time 
have been proposed for efficient inference.
The executed parts construct an inference path for a specific input, which is controlled by gating functions \cite{teja2018hydranets,veit2018convolutional,wang2018skipnet} or gating polices \cite{figurnov2017spatially,jie2019anytime}. Different from these methods, the executed path for data-dependent inference is controlled by the classifiers with different computation cost.

\textbf{Knowledge distillation.}
Knowledge distillation can improve the performance of small student networks by transferring ``knowledge'' from a large teacher network. Knowledge is encoded as extra supervision from either class posterior probabilities~\cite{hinton2015distilling,zhao2020highlight} or intermediate feature~\cite{ba2014deep,romero2015fitnets,zagoruyko2017paying}. 
Distillation is used in multi-exit frameworks by transferring knowledge from the last branch classifier~\cite{li2019improved}, intermediate attention features~\cite{zhang2019scan}, or by averaging all branch classifiers~\cite{phuong2019distillation}. We use a variant of online ensemble-based distillation~\cite{lan2018knowledge}, where the ensemble comes from weighting each of the branch classifiers according to its stage.

\section{Multi-Exit Networks}
\subsection{Basics}
Consider a multi-exit network 
where classifiers 
$1$ to $M\!-\!1$ are intermediate branch classifiers that segue off the backbone and classifier $M$ is the original classifier at the end of the backbone.  Each part of the backbone between branches are stages, 
where stage $m$ refers to the backbone processing between branch classifiers $m\!-\!1$ and $m$.  
Similar to~\cite{zhang2019scan}, we add branches where feature maps change in spatial resolution, \eg on VGG-16~\cite{simonyan2014very}, after the second, third and fourth pooling layers. 
It means that branches are added after the end of each stage, except the first and final ones. A branch at the very first decrease in spatial size is not considered as features are too simple for classification, while the original classifier at final stage has been already existed in backbones.
 
When considered as an ensemble, the multi-exit network returns $M$ classification outputs, \ie,
\begin{equation}
\small
\label{eq1}
\begin{split}
\big[y_1, \Compactcdots, y_M\big] & = \big[f_1(Z_1,\theta_1),  \Compactcdots, f_M(Z_M,\theta_M)\big]  \\
 & = \big[f_1(g_1(x,\phi_1),\theta_1),  \Compactcdots, f_M(g_M(x,\phi_M),\theta_M)\big],
\end{split}
\end{equation}
where $x$ is the input and $y_i\in\mathbb{R}^K$ is the soft-max output of the $i^{\text{th}}$ classifier $f_i$ for a $K$-class problem. The intermediate features $Z_i\in\mathbb{R}^{c_i\times h_i\times w_i}$ are generated by a transformation $g_i$ 
between $x$ and their weights $\phi_i$. 
Intermediate features $Z_i, i=1,\cdots M$ are inputs to the $i^{\text{th}}$ branch classifier $f_i$ which has corresponding parameter $\theta_i$, and also are outputs from the $i^{\text{th}}$ intermediate feature extractor $g_i$ of backbones or pre-trained models with the corresponding parameter $\phi_i$. Note that $\phi_i \subset \phi_j$ if $i \leq j$, as preceding stages are common to all branches.  

\subsection{Training}
Usually, multi-exit branch classifiers are trained cooperatively by minimizing a joint loss~\cite{huang2017multi,lee2015deeply,teerapittayanon2016branchynet}:
\begin{equation}
\small
\label{eq:loss}
\mathcal{L}_{\text{joint}} = \sum\limits_{i=1}^{M}\text{CE}_{i}(y,y_i) = \sum\limits_{i=1}^{M}\text{CE}_{i}\big(y,f_i(g_i(x,\phi_i),\theta_i)\big),
\end{equation}
where $y$ is the ground-truth and  $\text{CE}_i(\cdot,\cdot)$ denotes the cross-entropy loss of the $i^{\text{th}}$ classifier. 

 \begin{figure*}[t]
\centering
    \includegraphics[scale = 0.31]{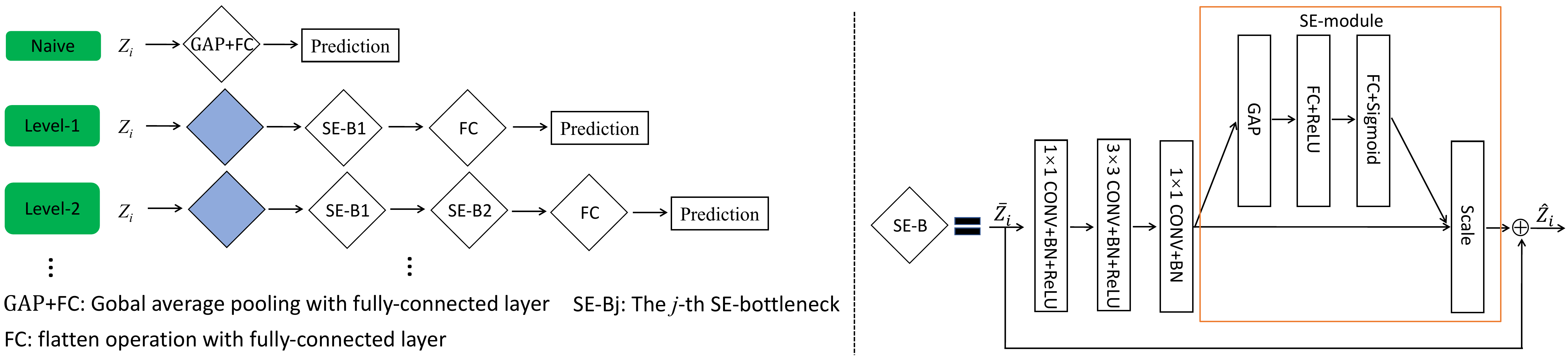}
\caption{Branch classifiers (left) and SE-B block architecture (right). A naive branch uses global average pooling with a fully-connected layer; other branch levels denote the number of SE-B blocks. The blue diamonds are convolutions. 
}
\label{fig:brancharch}
\end{figure*}

In addition, it has become standard to incorporate knowledge distillation. Here, the teacher used is either the end-stage classifier $f_M$~\cite{li2019improved,zhang2019scan}, or the mean prediction of all classifiers \cite{phuong2019distillation}
while students are the branch classifiers.  When knowledge distillation is incorporated, the loss becomes
\begin{equation}
\footnotesize
\label{eq_kd}
\mathcal{L_{\text{dist}}} = 
\lambda\mathcal{L}_{\text{joint}} + 
\tau^2(1-\lambda) \sum\limits_{i=1}^{M}\mathcal{L}_{\text{kl}}\big(\sigma(f'_i(Z_i,\theta_i)/\tau), \sigma(y_t/\tau)\big),
\end{equation}
where $\mathcal{L}_{kl}(\cdot, \cdot)$ denotes the Kullback-Leibler (KL) divergence. $f'_i$ and $y_t$ are the output of $f_i$ and the teacher's prediction before the soft-max operation $\sigma$, respectively. $\tau$ is a temperature parameter to soften the teacher output, and $\lambda$ is a weighting hyperparameter. In our experiments, we use $\tau=1$ and $\lambda=0.9$.  
As we wish to leverage the ensemble effect but also consider the strength of the individual classifiers in the ensemble, we propose a weighted ensemble, \ie 
$y_t = \sum_{j=1}^{M}\alpha_jf'_j(g_j(x,W_j),\theta_j)$, where $\alpha_j=j/\sum_{j=1}^{M}j$.

During training, we minimize Eq.~\ref{eq_kd} 
via stochastic gradient descent. Backbones and branch classifiers can be learned jointly from random initializations 
or one can leverage pre-trained networks as backbones. 

\begin{figure*}[t]
\centering
    \includegraphics[scale = 0.42]{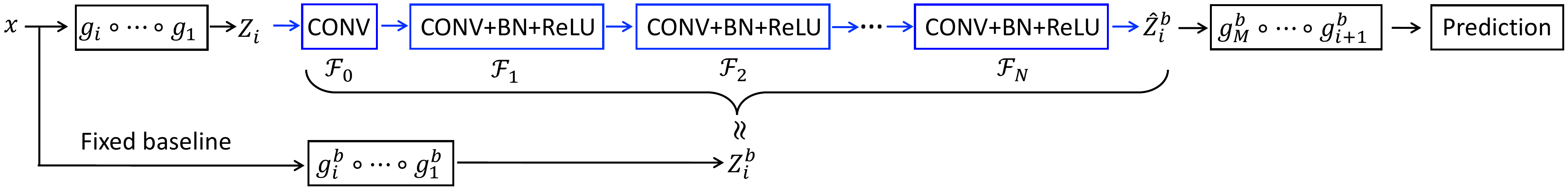}
\caption{Framework using knowledge consistency~\cite{liang2019knowledge} to  assessing branching impact at the $i^{\text{th}}$ stage.}
\label{fig:feature_complexity}
\end{figure*}

\subsection{Inference}
During inference, $y_i$ are estimated sequentially from $i\!=\!1$ to $M$, where the decision to estimate $y_{i+1}$ is based on $y_i$.  We opt for a simple thresholding of $y_i$; if $\max(y_i)\!<\!\gamma_i$, then we continue to estimate $y_{i+1}$ with classifier $f_{i+1}$ at the $(i+1)^{\text{th}}$ stage; otherwise the sample exits the classifier at the $i^{\text{th}}$ stage.  Previous works~\cite{jiang2018trust,nguyen2015deep} have shown that soft-max values are not always good indicators of classifier reliability.  Nevertheless, as these values are already computed, 
thresholding is still the favoured approach for determining branch exiting~\cite{huang2017multi,li2019improved,phuong2019distillation}.  Our branch classifiers are able to separate correct versus incorrectly labels only to some extent. 
Better measures of model confidence~\cite{corbiere2019addressing} could be substituted, but their utility is often limited by extra computational overhead which negate the gains afforded by dynamic inference.




\begin{table}[t]
\centering
\caption{Multi-exit architectures of ResNet-18 on ImageNet. $[a, b]\times c$ denotes the number of $c$ sequential operations using $a$ (\eg convolution or SE-B) to generate the number of $b$ feature maps. $\rightarrow k\times k, d$ in the classifiers means to generate the number of $d$ feature maps with the size of $k\times k$.}
\scriptsize
\begin{tabular}{|c|c|c|c|c}
\hline
	Layer & Output & ResNet-18 \\  
\hline
	Conv1 & $112\times 112$ & $7\times7$, 64, stride 2 \\ 
\hline
	Pooling & $56\times 56$  & $3\times3$ max pool, stride 2 \\
\hline
	Stage1 & $56\times 56$  & 
	$ \left[ \begin{array}{cc} 3\times3,64  \\     3\times3,64 \end{array}\right] \times  2$ \\ 
\hline
	\multirow{4}{*}{Classifier1} & Naive: 1000-d & GAP 1000-d FC, softmax \\
\cline{2-3}
	& Level-1: 1000-d & $\begin{matrix} [6\times6,256] \rightarrow 14\times14, 256 \\ [\text{SE-B},512] \rightarrow 7\times7, 512 \\ \text{1000-d, FC, softmax} \end{matrix}$ \\
\cline{2-3}
	& Level-2: 1000-d & $\begin{matrix} [6\times6,256] \rightarrow 14\times14, 256 \\ [\text{SE-B},512]\times2 \rightarrow 7\times7, 512 \\ \text{1000-d, FC, softmax} \end{matrix}$  \\
\cline{2-3}
	& Level-3: 1000-d & $\begin{matrix} [6\times6,256] \rightarrow 14\times14, 256 \\ [\text{SE-B},512]\times3 \rightarrow 7\times7, 512 \\ \text{1000-d, FC, softmax} \end{matrix}$  \\
\cline{2-3}
	& Level-4: 1000-d & $\begin{matrix} [6\times6,256] \rightarrow 14\times14, 256 \\ [\text{SE-B},512]\times4 \rightarrow 7\times7, 512 \\ \text{1000-d, FC, softmax} \end{matrix}$  \\
\hline
Stage2 & $28\times 28$  & 
	$ \left[ \begin{array}{cc} 3\times3,128  \\ 3\times3,128 \end{array}\right] \times  2$ \\  
\hline
\multirow{4}{*}{Classifier2} & Naive: 1000-d & GAP 1000-d FC, softmax \\
\cline{2-3}
	& Level-1: 1000-d & $\begin{matrix} [4\times4,256] \rightarrow 14\times14, 256 \\ [\text{SE-B},512] \rightarrow 7\times7, 512 \\ \text{1000-d, FC, softmax} \end{matrix}$   \\
\cline{2-3}
	& Level-2: 1000-d & $\begin{matrix} [4\times4,256] \rightarrow 14\times14, 256 \\ [\text{SE-B},512]\times2 \rightarrow 7\times7, 512 \\ \text{1000-d, FC, softmax} \end{matrix}$  \\
\cline{2-3}
	& Level-3: 1000-d & $\begin{matrix} [4\times4,256] \rightarrow 14\times14, 256 \\ [\text{SE-B},512]\times3 \rightarrow 7\times7, 512 \\ \text{1000-d, FC, softmax} \end{matrix}$   \\
\cline{2-3}
	& Level-4: 1000-d & $\begin{matrix} [4\times4,256] \rightarrow 14\times14, 256 \\ [\text{SE-B},512]\times4 \rightarrow 7\times7, 512 \\ \text{1000-d, FC, softmax} \end{matrix}$   \\
\hline
Stage3 & $14\times 14$  & 
	$ \left[ \begin{array}{cc} 3\times3,256  \\ 3\times3,256 \end{array}\right] \times  2$ \\  
\hline
\multirow{4}{*}{Classifier3} & Naive: 1000-d & GAP 1000-d FC, softmax \\
\cline{2-3}
	& Level-1: 1000-d & $\begin{matrix} [3\times3,256] \rightarrow 14\times14, 256 \\ [\text{SE-B},512] \rightarrow 7\times7, 512 \\ \text{1000-d, FC, softmax} \end{matrix}$  \\
\cline{2-3}
	& Level-2: 1000-d & $\begin{matrix} [3\times3,256] \rightarrow 14\times14, 256 \\ [\text{SE-B},512]\times2 \rightarrow 7\times7, 512 \\ \text{1000-d, FC, softmax} \end{matrix}$   \\
\cline{2-3}
	& Level-3: 1000-d & $\begin{matrix} [3\times3,256] \rightarrow 14\times14, 256 \\ [\text{SE-B},512]\times3 \rightarrow 7\times7, 512 \\ \text{1000-d, FC, softmax} \end{matrix}$   \\
\cline{2-3}
	& Level-4: 1000-d & $\begin{matrix} [3\times3,256] \rightarrow 14\times14, 256 \\ [\text{SE-B},512]\times4 \rightarrow 7\times7, 512 \\ \text{1000-d, FC, softmax} \end{matrix}$   \\
\hline
Stage4 & $7\times 7$  & 
	$ \left[ \begin{array}{cc} 3\times3,512  \\ 3\times3,512 \end{array}\right] \times  2$  \\
\hline
      Final Classifier & $1\times 1$   & GAP 1000-d FC, softmax \\
\hline
\end{tabular}
\label{tab_resnet_18_34_c}
\end{table}

\section{Branch  Classifier Design}
\subsection{SE-B Building Block}
\label{classifier}
In this work, we tackle the question of branch architecture, specifically the trade-off between branch depth and classification accuracy. The number of branches and placement have been explored previously in~\cite{teerapittayanon2016branchynet}.  In our branches, we use a basic building block of a bottleneck structure and a squeeze-excitation module. Bottleneck structures~\cite{he2016deep} are part of many modern CNNs and offer an excellent accuracy-cost trade-off.  Similarly, squeeze-excitation (SE) modules~\cite{hu2018squeeze} are a cheap
form of self-attention that help networks improve feature discriminability. We call this block the SE-bottleneck (SE-B, see Fig.~\ref{fig:brancharch} right) and vary the number of blocks to adjust branch complexity. For example, a single block in ResNet-18 requires about 13.7M FLOPs and is highly efficient.

\subsection{Branching Patterns}
For convenience of discussion and comparison, we equate branch complexity as the number of composing SE-B blocks and refer to this as the branch~\emph{level}.  Lower level branches are simpler and have less blocks, while higher level branches are more complex and have more blocks (see Fig.~\ref{fig:brancharch} left).  We investigate three branching patterns: \emph{constant-complexity} branching have branches all of the same level, \emph{complexity-increasing} branching has branch levels that increase with the stage number, and \emph{complexity-decreasing} branching has branch levels the decrease with the stage number.  

Branch architectures are difficult to compare on an absolute scale since one can always improve performance with higher level branches. We compare instead branching patterns while fixing the total branch levels, \ie the FLOP budget over all branches.  
Our branches are designed such that their 
cost depends only on the number of SE-B blocks regardless of the stage at which they are applied. This is achieved via a variable convolution layer (blue diamonds in Fig.~\ref{fig:brancharch}) to reduce all incoming $Z_i$ to the same size as $Z_{M-1}$. 
$Z_{M-1}$ itself undergoes a simple $3\times3$ convolution with stride 1 and padding 1 and has no change in resolution.
%
For example if $\{Z_1, Z_2, Z_3\}$ have 64, 128 and 256 channels respectively, the first convolution layer would be set with kernels of $6\!\times\!6$, $4\!\times\!4$ and $3\!\times\!3$ \footnote{As  $6\!\times\!6\!\times\!64 \approx 4\!\times\!4\!\times\!128$ and $6\!\times\!6\!\times\!64 = 3\!\times\!3\!\times\!256$.}. Following the convolution, the first SE-B reduces the spatial size of the branch input to half of the height and width while increasing the number of input channels by a factor of 2.  Subsequent SE-B blocks keep feature map size and the number of channels constant. We take ResNet-18 for example, detailed definitions of the branching architectures are given in Tables~\ref{tab_resnet_18_34_c}.

For comparison, we also consider a \emph{naive} branch architecture inspired by ResNet~\cite{he2016deep}, GoogLeNet~\cite{szegedy2015going} and DenseNet~\cite{huang2017densely}. The (end) classifier of all three architectures use global average pooling and a fully-connected layer; we use the same configuration in the naive branch.
However, multi-exit networks with naive branch adding into the intermediate classifiers achieves ultra-low performance (see Section~\ref{naive}), which is due to the disruption of semantic features whatever put in early or late state.

\subsection{Assessing Branching Impact}
\label{sec:consistency}
Methods which visualize CNN features~\cite{yosinski2015understanding,zeiler2014visualizing} have shown that visual concepts are encoded hierarchically over the course of a classifier, from simple to complex. We posit that for multi-exit classifiers to work successfully, the abstraction hierarchy of these visual concepts must be well aligned across all the classifiers, especially in the shared backbone. Adding branches is likely to disrupt the abstraction hierarchy. Directly quantifying the disruption is non-trivial so we propose the use of~\emph{knowledge consistency}~\cite{liang2019knowledge} indirectly probe the disruption.  

Knowledge is defined loosely by~\cite{liang2019knowledge} as the visual concepts encoded by intermediate layer features.  
The more similar the concepts for two networks, the more consistent the knowledge. Specifically, for two features $Z^p$ and $Z^q$ derived from input $x$, the consistent parts of $Z^q$ can be reconstructed from $Z^p$ since they both stem from $x$: 
\begin{equation}\label{eq:featrecon}
Z^q =\hat{Z}^q + \epsilon^q  
= \mathcal{F}_N\circ\cdots\circ\mathcal{F}_1\circ\mathcal{F}_0(Z^p) + \epsilon^q (N),
\end{equation}

\noindent where $\mathcal{F}_0$ is a linear transformation, $\{\mathcal{F}_1,\dots,\mathcal{F}_N$\} are non-linear transformations, and $\epsilon^q$ is the non-consistent part. For simplicity, we use one convolutional layer for $\mathcal{F}_0$ and a convolutional layer with batch normalization 
and a ReLU 
for each non-linear $\mathcal{F}_1,\dots,\mathcal{F}_N$.  $\{\mathcal{F}\}$ can be learned by minimizing the MSE between $\hat{Z}^q$ and $Z^q$ over all data samples. If ${Z}^p$ and ${Z}^q$ are sufficiently consistent, 
then the reconstructed $\hat{Z}^q$ can be used to substitute $Z^q$.  However, in Eq.~\ref{eq:featrecon}, $\epsilon_q$ can be adjusted by $N$; presumably, the higher $N$, the smaller the $\epsilon_q$. As such, one needs to interpret results as a whole over multiple $N$, or ``fuzziness levels'' as per~\cite{liang2019knowledge}.

 
We further define $Z^b_i = g_{i}^{b}\circ\cdots\circ g_1^b (x)$ as feature outputs of stage $i$ from a backbone baseline with no branches, versus corresponding $Z_i = g_i\circ\cdots\circ g_1 (x)$ when branch classifiers are appended.  To check for disruptions, the reconstructed $\hat{Z}^b_i=\mathcal{F}_N \circ \cdots \circ \mathcal{F}_0 (Z_i)$ are substituted for $Z^b_i$ before applying the remainder of the original baseline $g_M^b\circ\cdots\circ g_{i+1}^b (\hat{Z}^b_i)$ (see Fig.~\ref{fig:feature_complexity}). We compare the final accuracy using $\hat{Z}^b_i$ over different $N$ to check for consistency. Branched networks with features more aligned hierarchically with the original baseline will have $Z_i$ that are more consistent with $Z^b_i$.  Networks whose feature are less aligned, \ie more disrupted in hierarchy, will have features less consistent and hence yield lower classification accuracies with the reconstructed $\hat{Z}^b_i$.

\section{Experiments}\label{sec:experiments}

\begin{figure*}[t]
\centering
  \subfigure[Level:n+2+3]{
    \includegraphics[scale = 0.275]{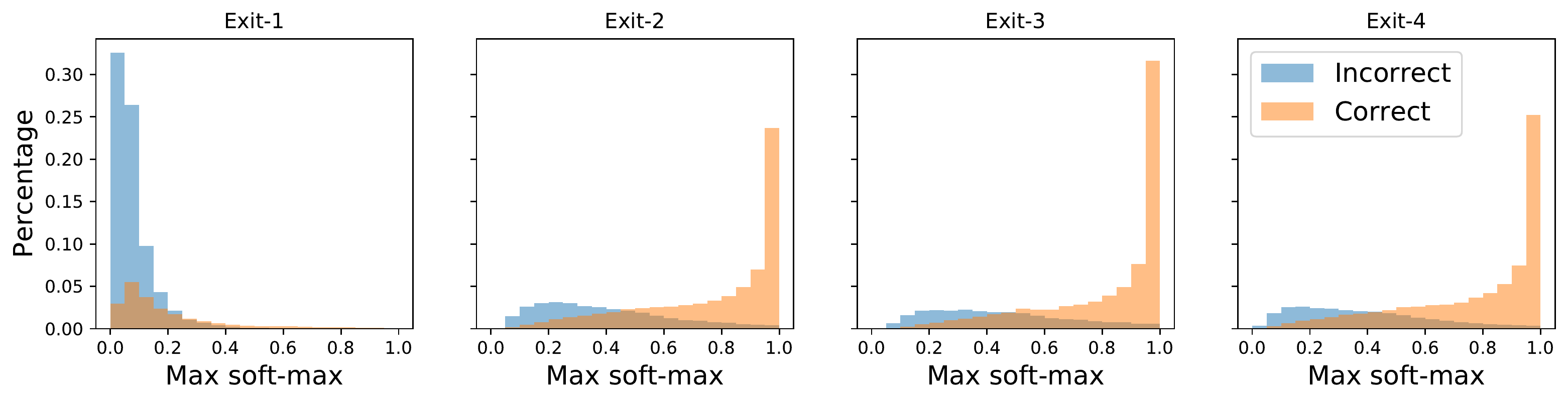}
    \label{fig_confidence_1}
  }
  \subfigure[Level:1+2+3]{
    \includegraphics[scale = 0.275]{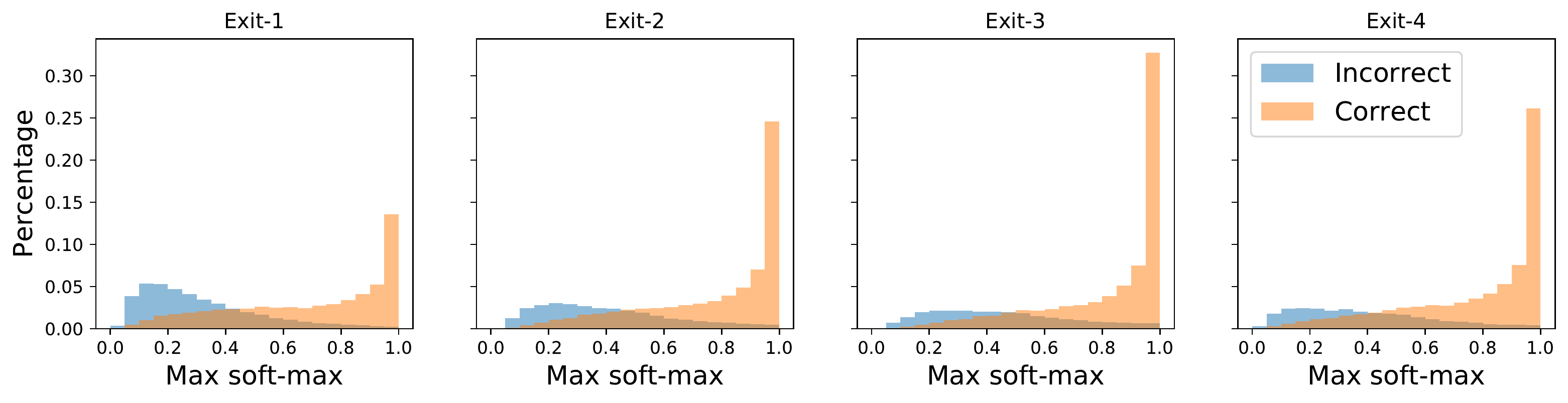}
    \label{fig_confidence_2}
  }
\subfigure[Level:n+n+n]{
    \includegraphics[scale = 0.275]{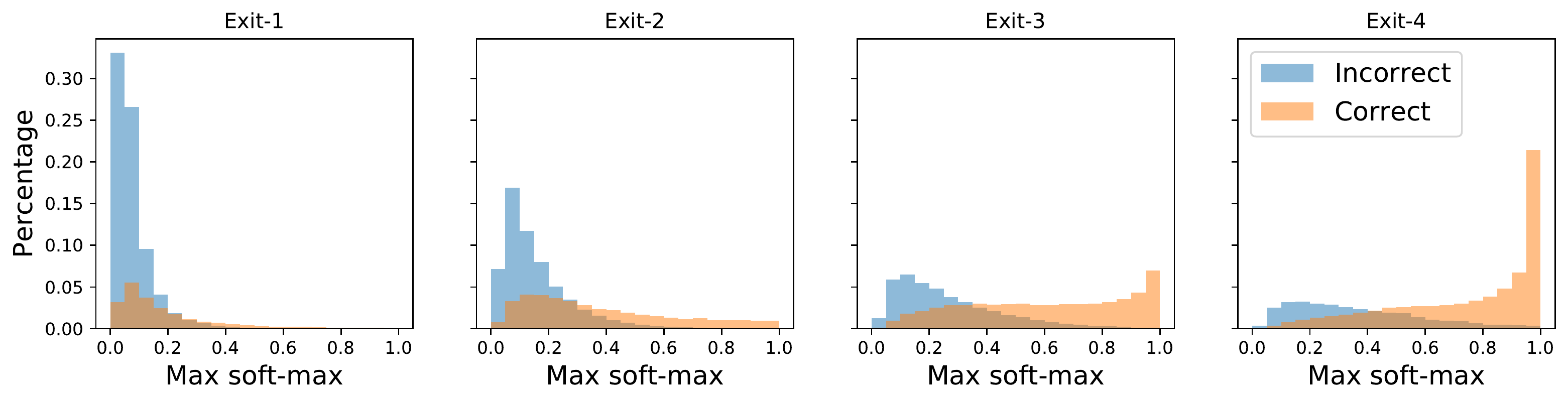}
    \label{fig_confidence_3}
  }
  \subfigure[Level:2+2+2]{
    \includegraphics[scale = 0.275]{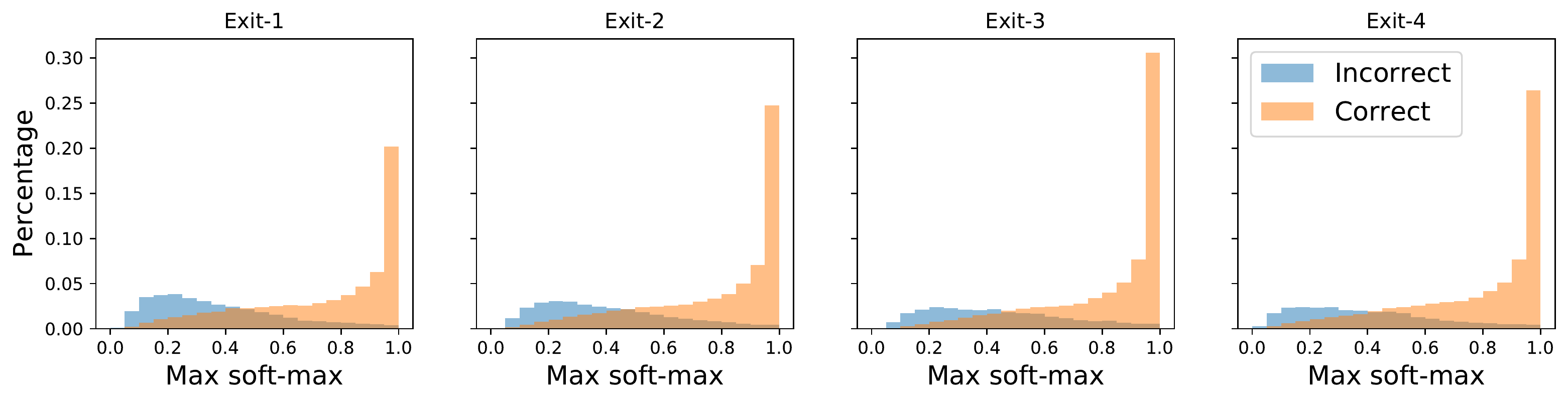}
    \label{fig_confidence_4}
  }
  \subfigure[Level:3+2+n]{
    \includegraphics[scale = 0.275]{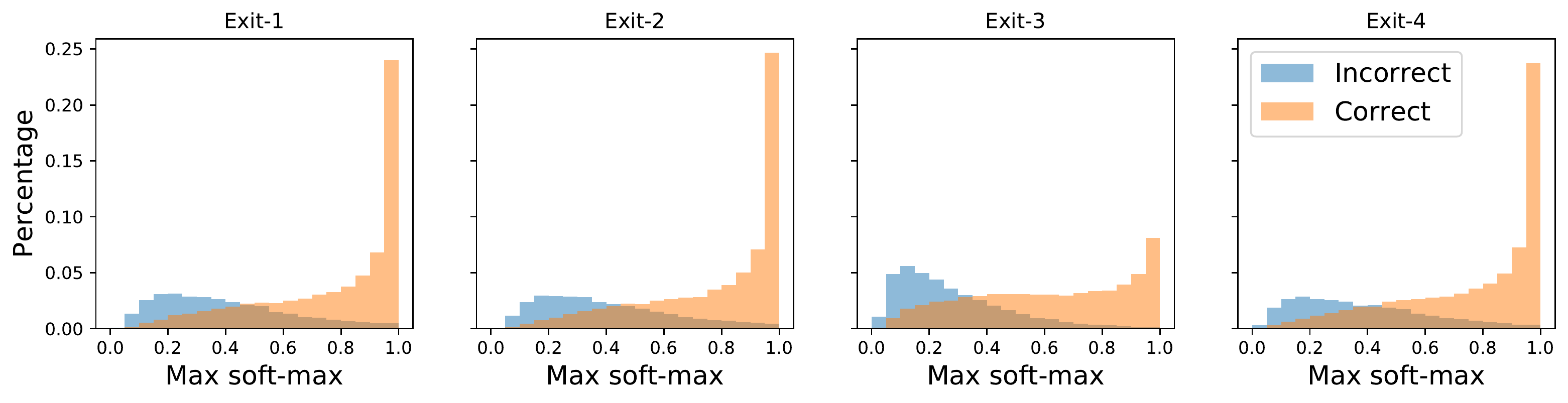}
    \label{fig_confidence_5}
  }
  \subfigure[Level:3+2+1]{
    \includegraphics[scale = 0.275]{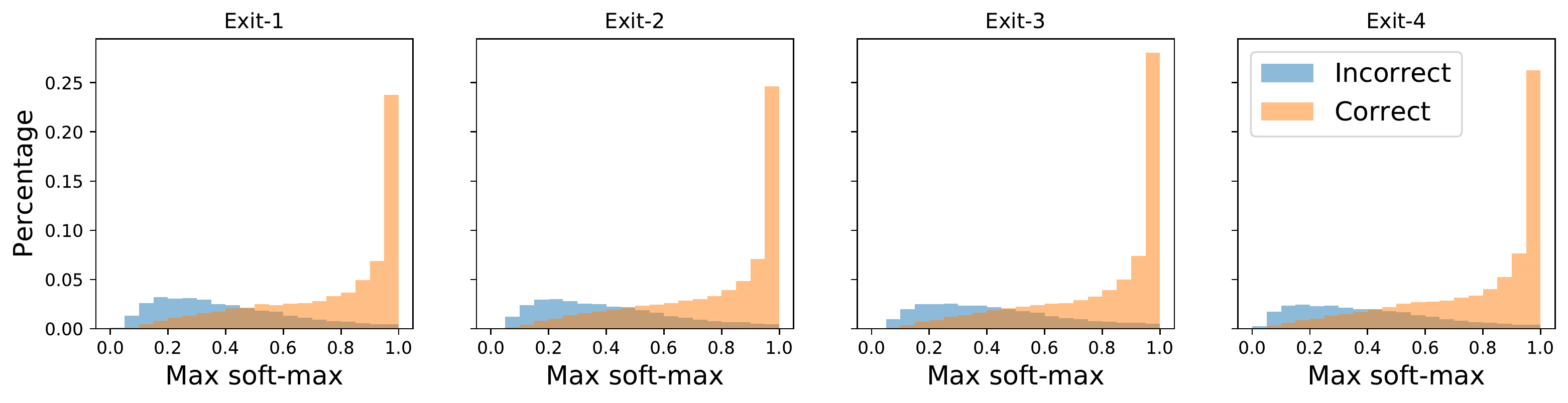}
    \label{fig_confidence_6}
    }
 \caption{Histogram of max soft-max values and percentage of incorrect/correct sample rates using different combinations of branch classifiers on the validation set of ResNet-18. n denotes a naive branch and the classifier in Exit-4 is the original final one. (a)(b), (c)(d), and (e)(f) are complexity-increasing, constant complexity and complexity-decreasing branching patterns respectively. Compared to SE-block branches of various levels, the naive branch classifiers have both lower max soft-max values and poor separability between the incorrect and incorrect samples. 
 }
 \label{fig_confidence}
\end{figure*}

\subsection{Experimental Setup}
\textbf{Backbones and Datasets.} We consider as backbones single-path networks (\eg VGG~\cite{simonyan2014very}) and multi-path networks (\eg GoogLeNet~\cite{szegedy2015going}, ResNet~\cite{he2016deep}, DenseNet~\cite{huang2017densely} and MobileNetV2 \cite{sandler2018mobilenetv2}). The task at hand is image classification, which we apply to the  CIFAR-100~\cite{krizhevsky2009learning} and ImageNet 2012~\cite{russakovsky2015imagenet} datasets.  CIFAR-100 contains 50k/10k $32\times32$ training/test images from 100 classes. 
We apply standard data augmentation methods~\cite{he2016deep} for training and hold out 5K training images as a validation set.  ImageNet 2012 has 1.28M training images over 1K classes and 50K validation images upon which we test. To validate hyper-parameters, we instead hold out 50K images (50 per class) from the training set.  
%
%
For data augmentation on ImageNet, the training images are rescaled to $256\times256$, with a $224\times224$ crop randomly sampled from each image and flipped horizontal.  Test accuracy is measured on the validation set using only the central $224\times224$ patch.

\textbf{Implementation.} All models are trained using stochastic gradient descent in PyTorch
on NVIDIA RTX 2080Ti GPUs. We train all multi-exit models with an initial learning rate of 0.01 and use momentum and a linear decay~\cite{li2019budgeted}.  The weight decay and momentum are set to 0.0001 and 0.9 respectively. For CIFAR-100, all the models are trained for 300 epochs in batches of 64. For ImageNet, training is done over 60 epochs in batches of 256, except MobileNetV2 with 90 epochs. Note that this training setting is not consistent to the standard ImageNet training protocols\footnote{https://github.com/pytorch/examples/tree/master/imagenet}, as the linear decay strategy is used to implement fast convergence by the training with smaller epochs.  

For assessing branching impact, we use the same training settings on ImageNet, but with 10 epochs\footnote{We use the smaller number of epoch than that in the normal multi-exit training, as this process acts like a simple fine-tuning for fast convergence} and an initial learning rate of 0.01 with 0.1 decay every 3-epoch. For stage-wise training, each stage and its associated branch are trained with the same number of epochs as the total epochs in cooperative learning. The total training epochs are therefore increased $M-$fold. For knowledge distillation, we set hyper-parameter $\lambda$ and $\tau$ to 0.9 and 1, respectively. Unless otherwise noted, we use the standard \emph{cooperative training} given in Eq.~\ref{eq:loss} without knowledge distillation. 

\textbf{Evaluation.} 
For performance evaluation, we report top-1 accuracy.  Accuracy-wise, we consider the individual branch classifiers ($f_i$) over all samples in the dataset as well as the Top-1 adaptive accuracy under the multi-exit setting, \ie by exiting samples according to the threshold $\gamma$. For computational cost, we report the total FLOPs (backbone plus all branch classifiers) for each classifier $f_i$ individually and the adaptive average FLOPS (AF). Like~\cite{huang2017multi,li2019improved}, the adaptive average is estimated as the weighted summation of the branches' based on the percentage of exited samples:
\begin{equation}
\small
\label{af}
AF = \Sigma_{m=1}^{M}r_m(\gamma_m) \Big(\text{FLOPs}(f_m)+\Sigma_{i=1}^{m-1}\big(\text{FLOPs}(f_i - B_i)\big)\Big),
\end{equation}

\noindent where $\text{FLOPs}(f_m)$ indicates the number of FLOPs for classifier $m$, which contains only one branch classifier and backbone at the $m^{\text{th}}$ stage.
$r_m$ is percentage of samples exiting at the $m^{\text{th}}$ stage.  Note that $r_m$ is controlled by a stage-specific threshold $\gamma_m$.  In the second summation term, $B_i$ is the backbone architecture of the $i^{\text{th}}$ stage; this term accounts for the earlier branch classifiers first traversed by the sample.

\begin{table*}[t]
\footnotesize
\caption{Comparison of branching patterns with ResNet-18 backbone on ImageNet.  $f_1$-$f_3$ are branch classifiers with complexity given by levels, while $f_4$ is the end-stage classifier from the backbone.  Baselines without branches (\ie ResNet-18 and VGG-16) have accuracies of  69.76\% and 73.36\% with 1.81B FLOPs and 15.47B FLOPs.  
}

\begin{center}
\scalebox{1}{
\bgroup
\def\arraystretch{1.15}
\begin{tabular}{|c|c|c|c|c|c|c|c|c||c|c|c|c|c|c|c|c|}
\hline
\multirow{3}*{Level} & \multicolumn{8}{c||}{ResNet-18} & \multicolumn{8}{c|}{VGG-16} \\ 
\cline{2-17}
& \multicolumn{4}{c|}{ Branch Classifier Accuracy (\%) } & \multicolumn{4}{c||}{ Branch Classifier FLOPS (B) } & \multicolumn{4}{c|}{ Branch Classifier Accuracy (\%) } & \multicolumn{4}{c|}{ Branch Classifier FLOPS (B) }\\
\cline{2-17}
 & $f_1$ & $f_2$ & $f_3$ & $f_4$ & $f_1$ & $f_2$ & $f_3$ & Total & $f_1$ & $f_2$ & $f_3$ & $f_4$ & $f_1$ & $f_2$ & $f_3$ & Total \\
\hline
1+2+3 &  55.34 & 65.43  & 69.95  & 69.11 & 0.74 & 1.17 & 1.62 & \multirow{3}*{2.37} & 62.04 & 71.84 & 74.30 & 73.07 & 5.32 & 10.00&  14.78 & \multirow{3}*{17.56} \\ 
2+2+2 & 60.70 & 65.43  & 69.19  & 69.26 & 0.77 & 1.17 & 1.59 & & 67.39 & 71.84 & 74.59 & 73.33 & 5.42 & 10.00 & 14.67 &  \\ 
3+2+1 & 63.91 & 65.34  & 67.64 & 68.96 & 0.80 & 1.17 & 1.57 & & 69.56 & 71.30 & 73.01 & 73.38 & 5.53 & 10.00 & 14.57 & \\ 
\hline
2+3+4 & 61.03 & 67.42  & 70.73  & 69.19 & 0.77 & 1.20 & 1.65 & \multirow{3}*{2.45} & 67.16 & 73.04 & 76.12 & 73.54 & 5.42 & 10.10 & 14.88 & \multirow{3}*{17.87} \\
3+3+3 & 64.08 & 67.05  & 70.01  & 69.40 & 0.80 & 1.20 & 1.62 & &  70.12 & 73.45 & 75.59 & 73.47 & 5.53 & 10.10 & 14.78 & \\
4+3+2 &  65.44 & 67.06  & 69.22  & 69.28 & 0.82 & 1.20 & 1.59 & & 71.34 & 73.26 & 74.86 & 73.36 & 5.64 & 10.10 & 14.67 & \\
\hline
\end{tabular}
\egroup
}
\label{tab_resnet18_vgg}
\end{center}
\end{table*}

\subsection{Naive Branch Classifiers}
\label{naive}
We first explore the baseline naive classifier as a branch classifier, using ResNet-18 backbone on ImageNet. 
We add three branch classifiers ($f_1$ to $f_3$) and organize our combination of classifiers into three patterns: complexity-increasing, constant-complexity and complexity-decreasing branching patterns. Note that $f_4$ (also denoted as Exit-4) is the original final classifier, including global average pooling and one fully-connected layer.

\begin{figure}[t]
\centering
  \subfigure[ResNet-18]{
    \includegraphics[scale = 0.205]{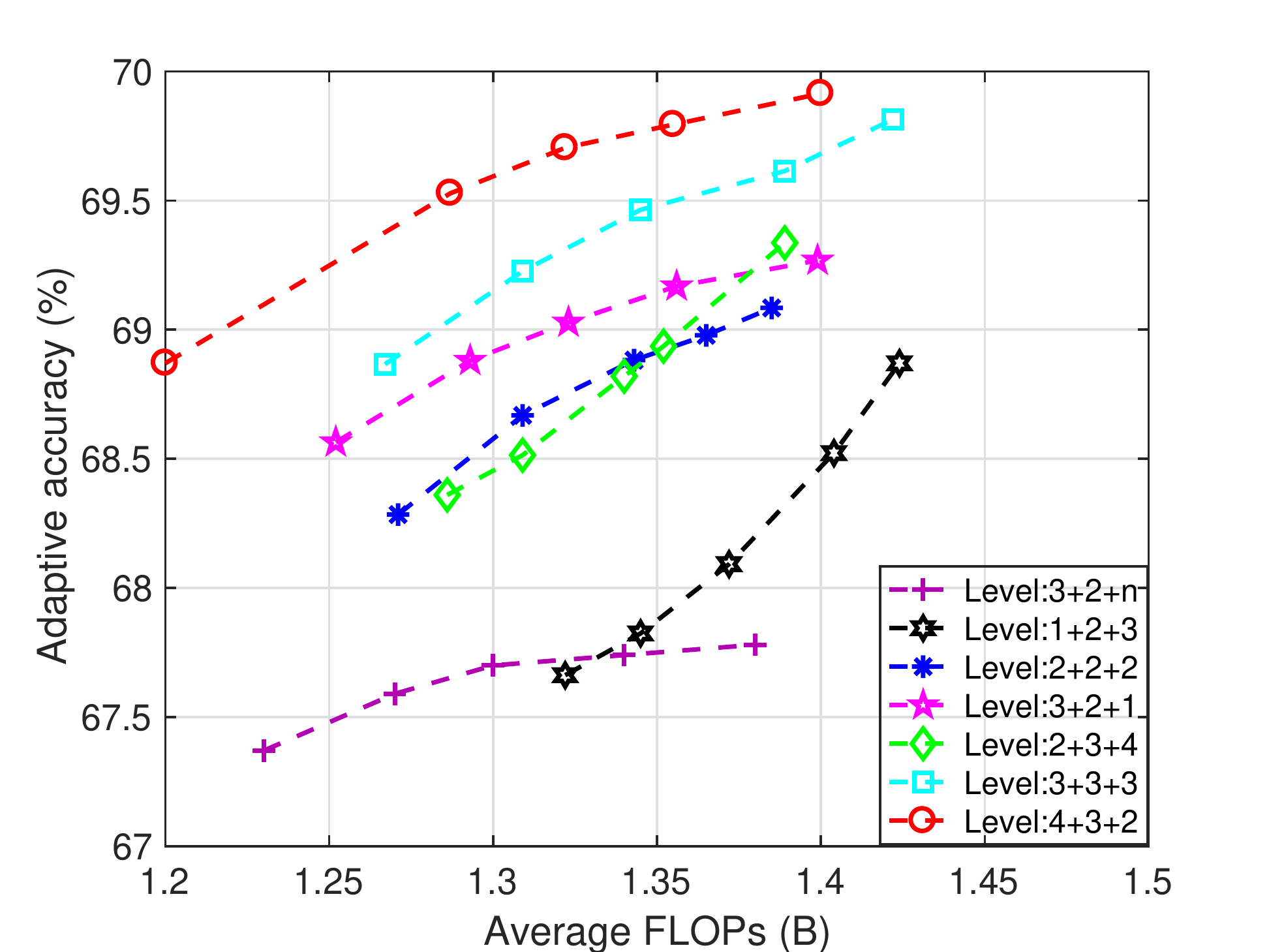}
    \label{fig_af_ac_a}
  }
  \subfigure[VGG-16]{
    \includegraphics[scale = 0.205]{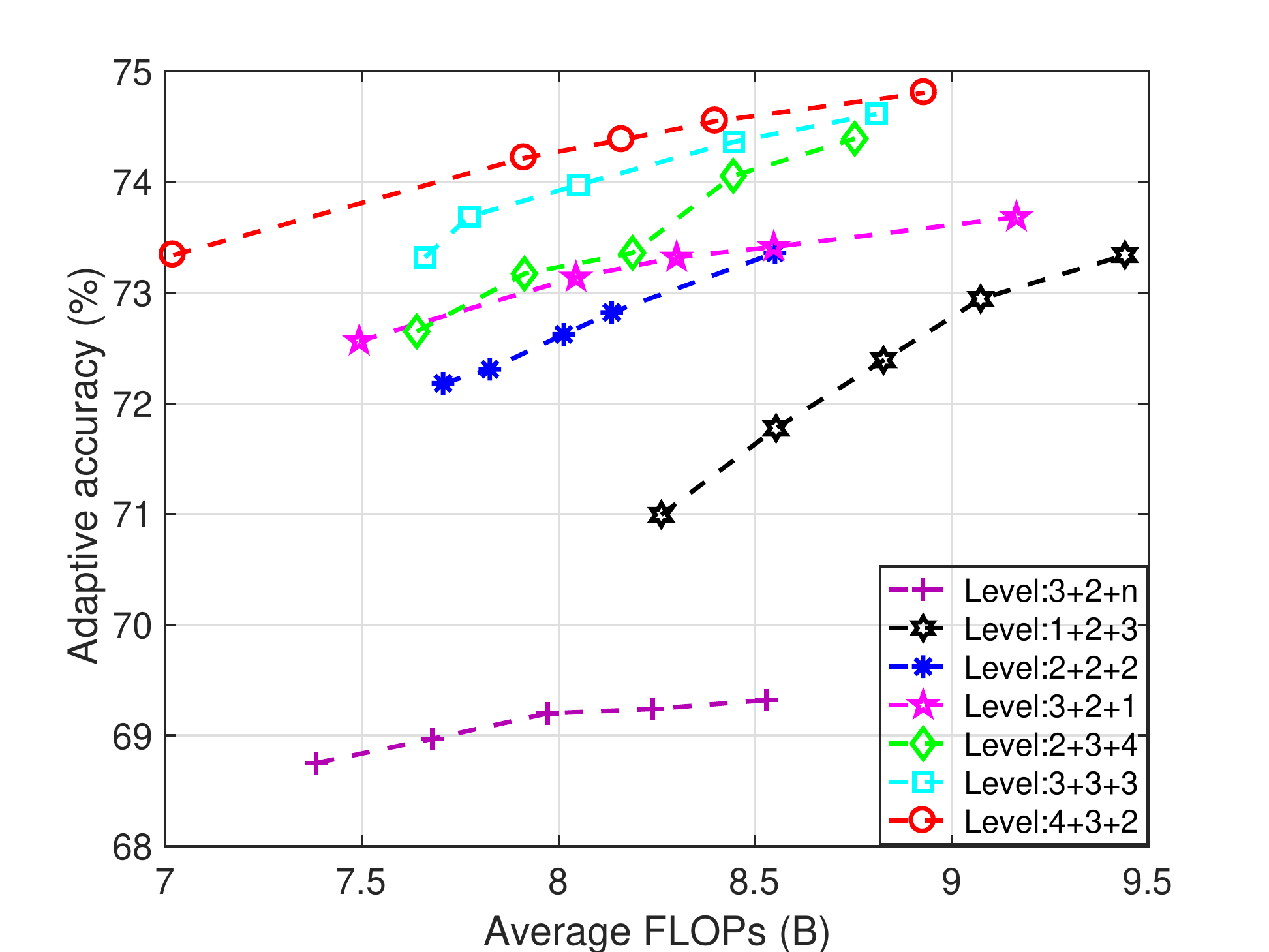}
    \label{fig_af_ac_b}
  }
\caption{Adaptive results of branching patterns for ResNet-18 and VGG-16 backbone on ImageNet. we can observe that (1) with the similar branching patterns, the trend of adaptive results keeps the relative consistent and increasing the number of total SE-B blocks improves the adaptive performance; (2) the complexity-decreasing branching patterns of 4+3+2 achieves the best adaptive results, compared to that on the complexity constant branching of 3+3+3 and the complexity-increasing branching of 2+3+4; (3) naive branch adding at the late stage achieves the worse adaptive results.}
\label{fig_af_ac}
\end{figure}

Fig.~\ref{fig_confidence} shows histograms of correctly versus incorrectly classified samples from the validation data based on the highest soft-max value.  The histograms show that softmax values for
naive branch classifiers in the early stages are much lower than other branch classifiers (\eg compare (a) n+2+3 and (c) n+n+n with (b) 1+2+3 and (d) 2+2+2 respectively).  Furthermore, the correct and incorrect sample distributions overlap for the naive classifiers, so that they are not separable.  Separation for the naive classifiers improve if they are used in later classifiers (\eg see (e) 3+2+n), but a relatively high $\gamma$ is still necessary, which leaves many samples to be processed by the final classifier and accumulates significant computational cost.  
This effect is reduced if the naive branch is used in later stages (\eg the third stage) in the complexity-decreasing pattern, as shown in Fig.~\ref{fig_confidence_5}. However, naive branches are still not good choice adding into backbones for early exiting, whose performance is presented in Fig.~\ref{fig_af_ac}. This is due to the disruption of semantic features when adding the naive classifier into the intermediate features. For other classifiers, they are able to reasonably separate correct and incorrectly samples. Therefore, we no longer discuss the naive classifiers in the remaining analysis for better analyzing the branching patterns under the constraint of the same total FLOPs.  

\subsection{Branching Patterns}
\label{bp}

\begin{figure*}[t]
\centering
    \subfigure[Samples exited rate on ResNet-18]{
    \includegraphics[scale = 0.28]{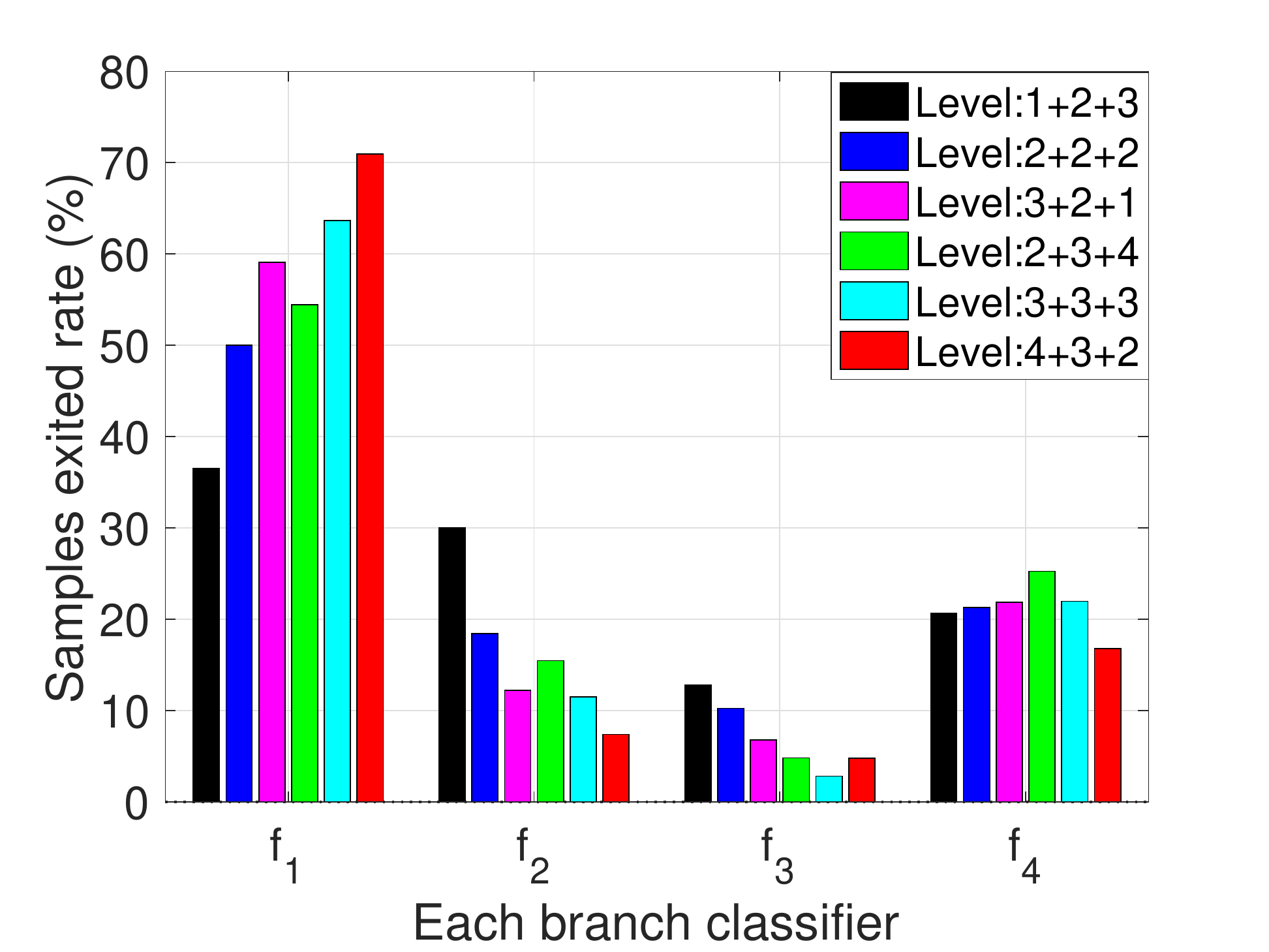}
    \label{fig_exit_a}
    }
    \subfigure[Samples exited rate on VGG-16]{
    \includegraphics[scale = 0.28]{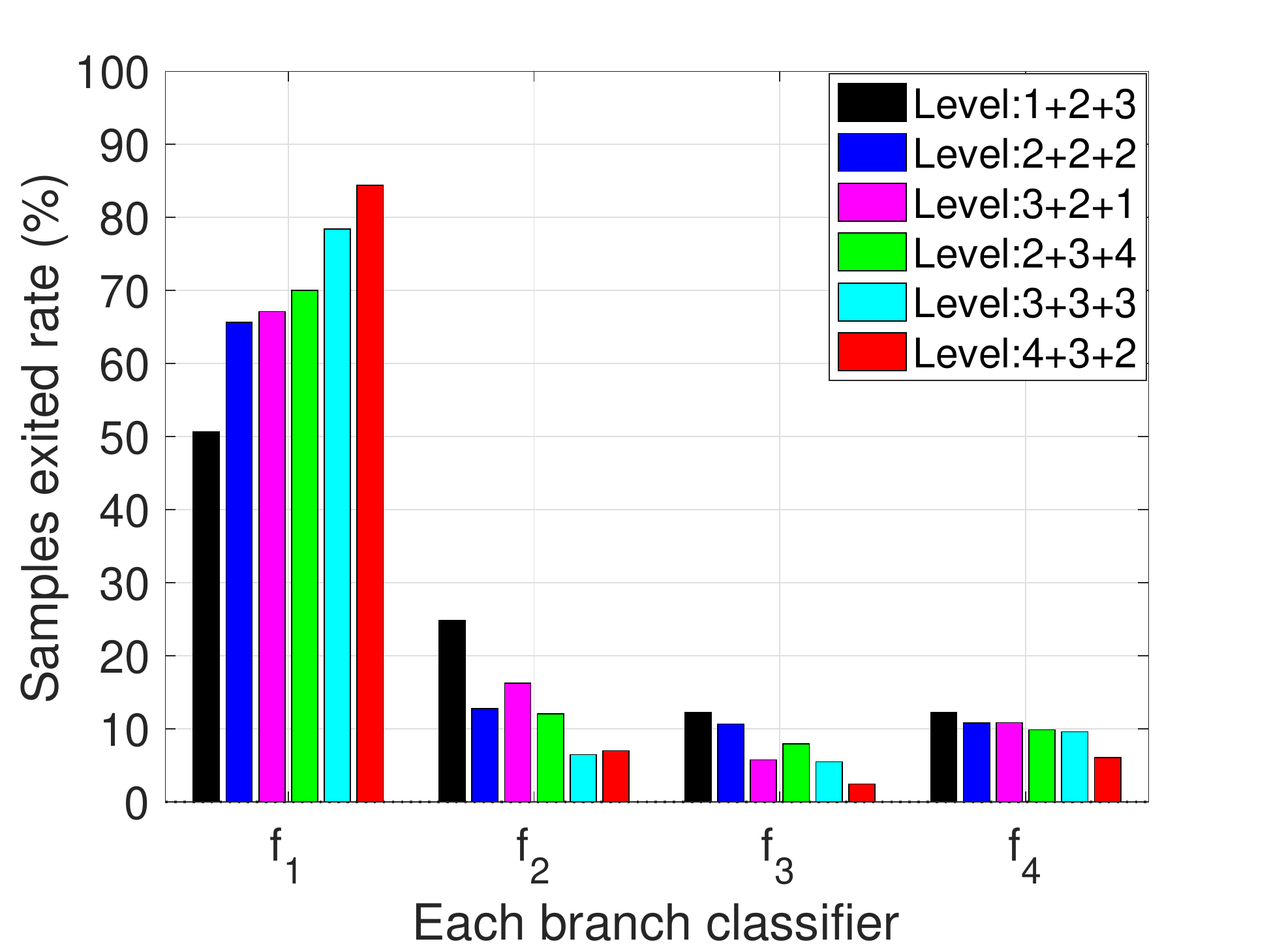}
    \label{fig_exit_b}
    }
    \subfigure[Average FLOPs on ResNet-18 with adaptive accuracy of about 68.88\%]{
    \includegraphics[scale = 0.28]{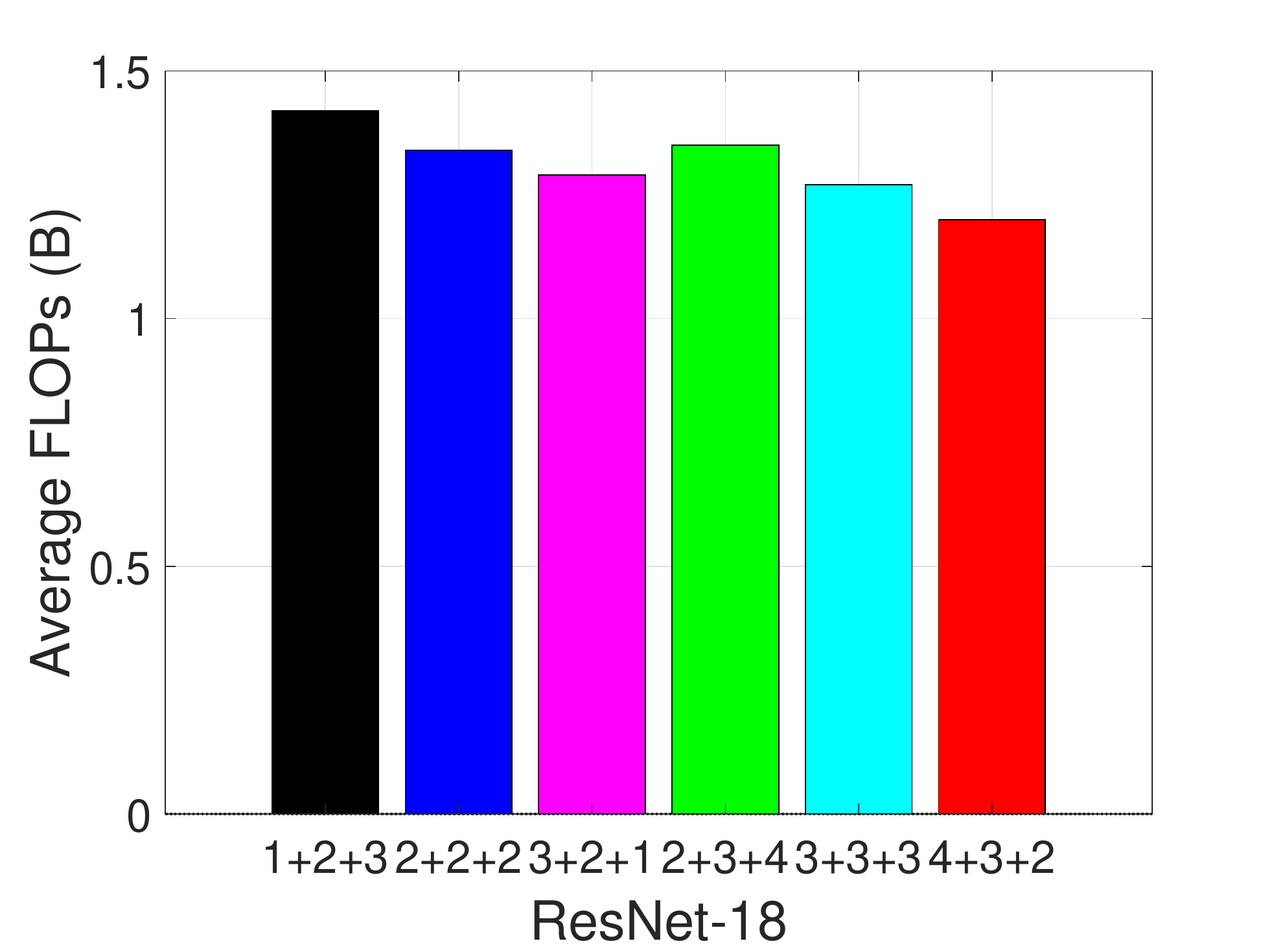}
    \label{fig_exit_c}
    }
    \subfigure[Average FLOPs on VGG-16 with adaptive accuracy of about 73.36\%]{
    \includegraphics[scale = 0.28]{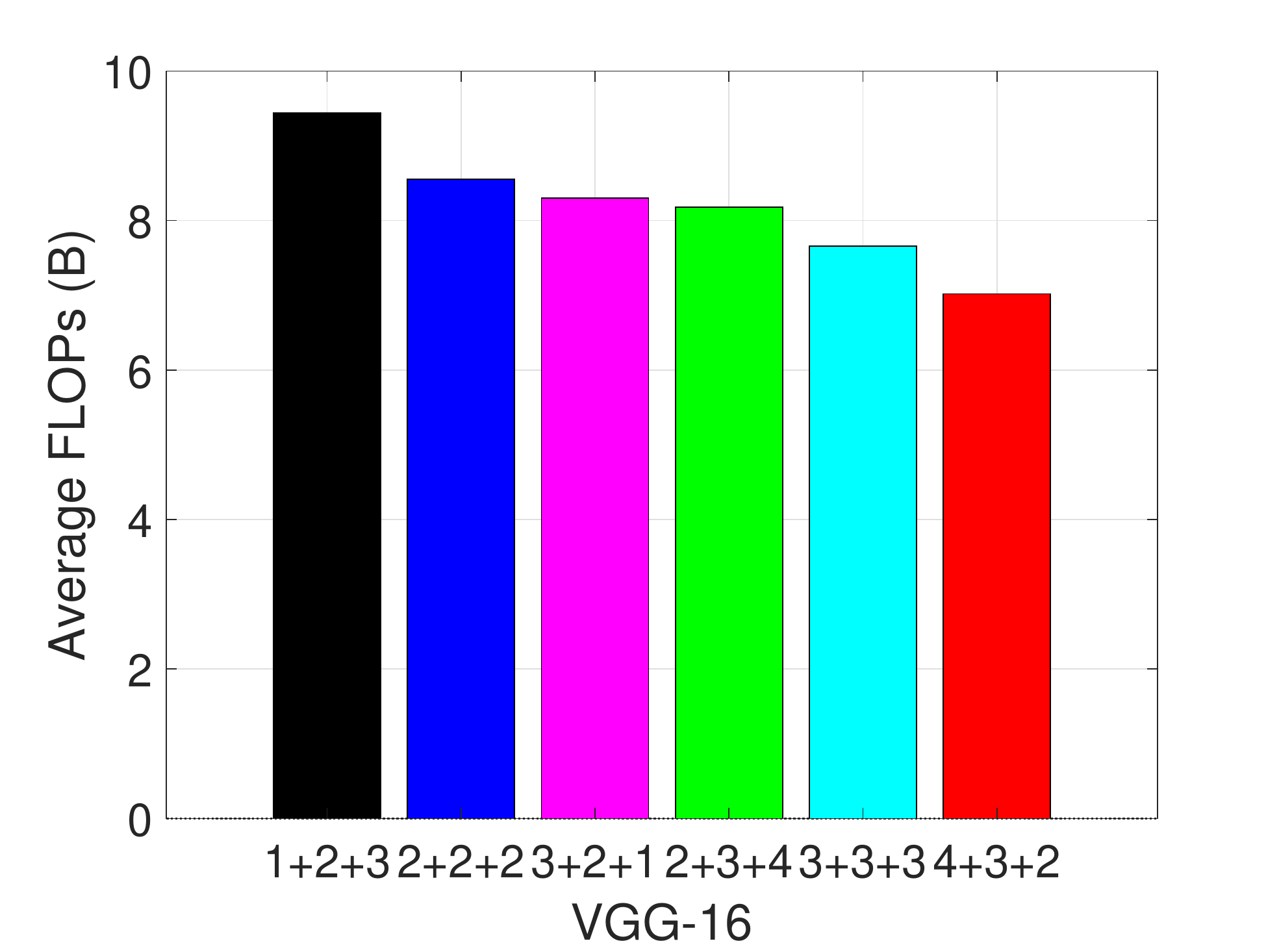}
    \label{fig_exit_d}
    }
    \subfigure[Thresholds on ResNet-18]{
    \includegraphics[scale = 0.28]{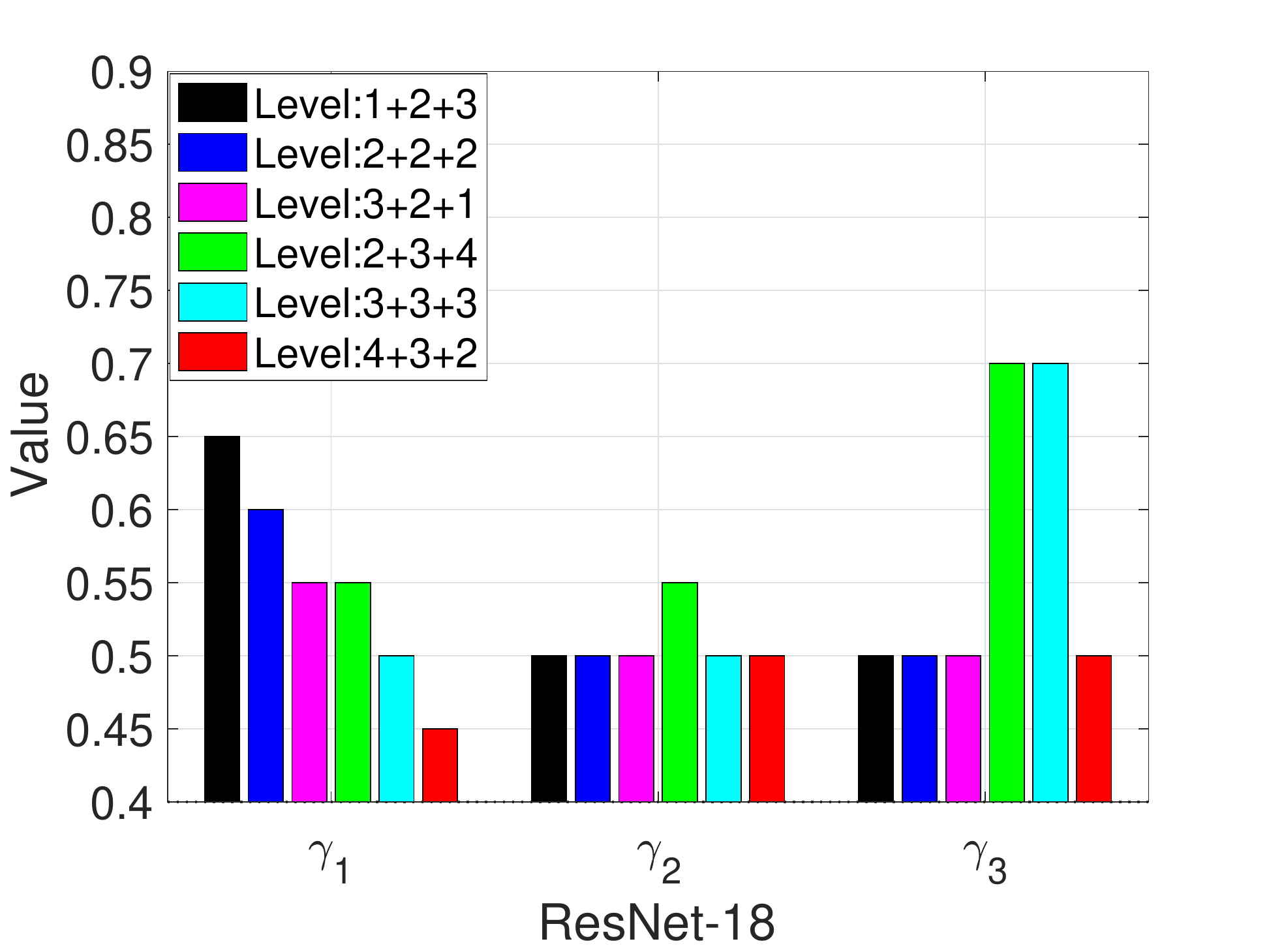}
    \label{fig_exit_e}
    }
    \subfigure[Thresholds on VGG-16]{
    \includegraphics[scale = 0.28]{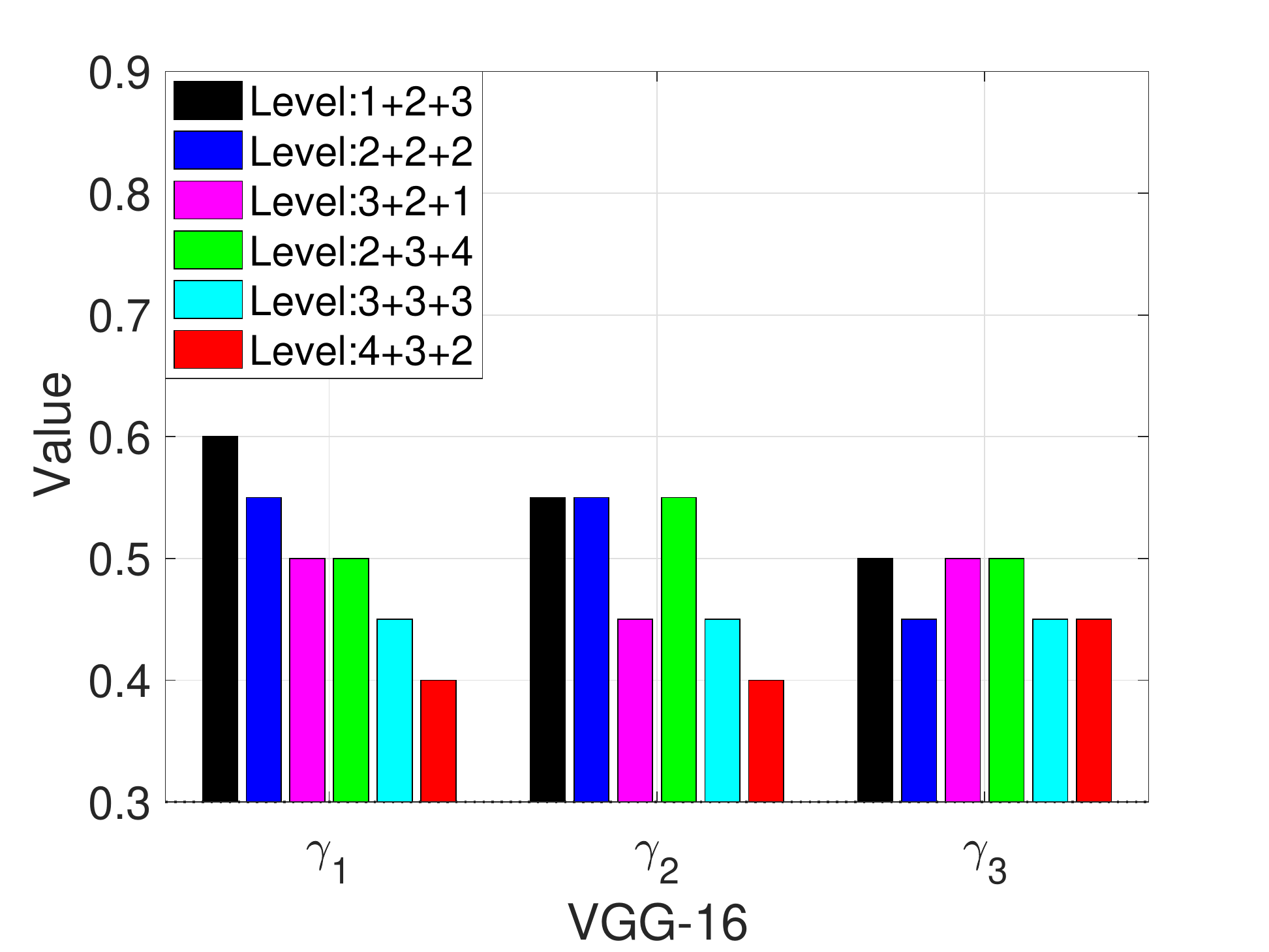}
    \label{fig_exit_f}
    }
\caption{Effect of branching patterns at the fixed adaptive accuracy on ResNet-18 and VGG-16. (a) and (b) are the percentages of samples exited at each branches on ResNet-18 and VGG-16, respectively. (c) and (d) are the corresponding average FLOPs of branch classifiers in (a) and (b) at the relative same adaptive accuracy, respectively. (e) and (f) are the corresponding threshold $\gamma$ for the former three stages to achieve (a) and (b), respectively.  Figure best viewed in colour.}
 \label{fig_sample_exit}
\end{figure*}

For fair comparison, one should consider only the configurations within the same total classifier FLOPs grouping. On ImageNet, we use two different total classifier FLOPs: 2.37B and 17.56B for 6 total SE-B blocks and 2.45B and 17.87B for 9 total SE-B blocks on ResNet-18 and VGG-16, respectively. On CIFAR-100, we use the total classifier FLOPs of 132.8M, 260.2M, 443.18M and 1.52B for 4 total SE-B blocks on ResNet-56, ResNet-110, DenseNet-40-12 and GoogLeNet, respectively. These blocks are organized into complexity increasing, constant-complexity and complexity-decreasing branching patterns. 

\begin{table*}[t]
\footnotesize
\caption{Comparison of branching patterns with ResNet-56 and ResNet-110 backbones on CIFAR-100. Branchless baselines of ResNet-56, ResNet-110, DenseNet-40-12 and GoogLeNet have accuracies of 71.65\%, 71.94\%, 73.44\% and 78.39\% with 125.49M FLOPs, 252.89M FLOPs, 282.96M FLOPs and 1.52B, respectively.}
\begin{center}
\scalebox{1}{
\bgroup
\def\arraystretch{1.15}
\begin{tabular}{|c|c|c|c|c|c|c||c|c|c|c|c|c|}
\hline
\multirow{3}*{Level} & \multicolumn{6}{c||}{ResNet-56} & \multicolumn{6}{c|}{ResNet-110} \\ 
\cline{2-13}
& \multicolumn{3}{c|}{ Branch Classifier Accuracy (\%) } & \multicolumn{3}{c||}{ Branch Classifier FLOPS (M) } & \multicolumn{3}{c|}{ Branch Classifier Accuracy (\%) } & \multicolumn{3}{c|}{ Branch Classifier FLOPS (M) } \\
\cline{2-13}
 & $f_1$ & $f_2$ & $f_3$ & $f_1$ & $f_2$ & Total & $f_1$ & $f_2$ & $f_3$ &  $f_1$ & $f_2$ & Total\\
\hline
1+3 & 59.87 & 69.20 & 73.02 & 45.89 & 88.52 & \multirow{3}*{132.80} & 61.53 & 70.03 & 73.65 & 88.36 & 173.46 & \multirow{3}*{260.20} \\
2+2 & 64.46 & 69.26 & 72.16 & 46.43 & 87.98 &  & 64.96 & 70.12 & 73.70 & 88.90 & 172.92 & \\ 
3+1 & 64.52 & 69.32 & 72.80 & 46.97 & 87.44 &  & 65.14 & 70.81 & 74.22 & 89.44 & 172.38 & \\
\hline
\hline
\multirow{3}*{Level} & \multicolumn{6}{c||}{DenseNet-40-12} & \multicolumn{6}{c|}{GoogLeNet} \\ 
\cline{2-13}
& \multicolumn{3}{c|}{ Branch Classifier Accuracy (\%) } & \multicolumn{3}{c||}{ Branch Classifier FLOPS (M) } & \multicolumn{3}{c|}{ Branch Classifier Accuracy (\%) } & \multicolumn{3}{c|}{ Branch Classifier FLOPS (B) } \\
\cline{2-13}
 & $f_1$ & $f_2$ & $f_3$ & $f_1$ & $f_2$ & Total & $f_1$ & $f_2$ & $f_3$ &  $f_1$ & $f_2$ & Total\\
\hline
1+3 & 73.20 & 76.45 & 73.05 & 215.02 & 345.77 & \multirow{3}*{443.18} & 76.49 & 80.65 & 79.65 & 1.10 & 2.02 & \multirow{3}*{2.68} \\
2+2 & 72.90 & 76.35 & 72.18 & 227.92 & 332.87 &  & 76.47 & 79.92 & 78.89 & 1.19 & 1.93 & \\ 
3+1 & 73.84 & 76.24 & 72.85 & 240.81 & 319.97 &  & 77.30 & 80.01 & 78.97 & 1.28 & 1.84 & \\
\hline
\end{tabular}
\egroup
}
\end{center}
\label{tab_resnet56_110}
\end{table*}

\begin{figure*}[t]
\centering
  \subfigure[ResNet-56]{
    \includegraphics[scale = 0.21]{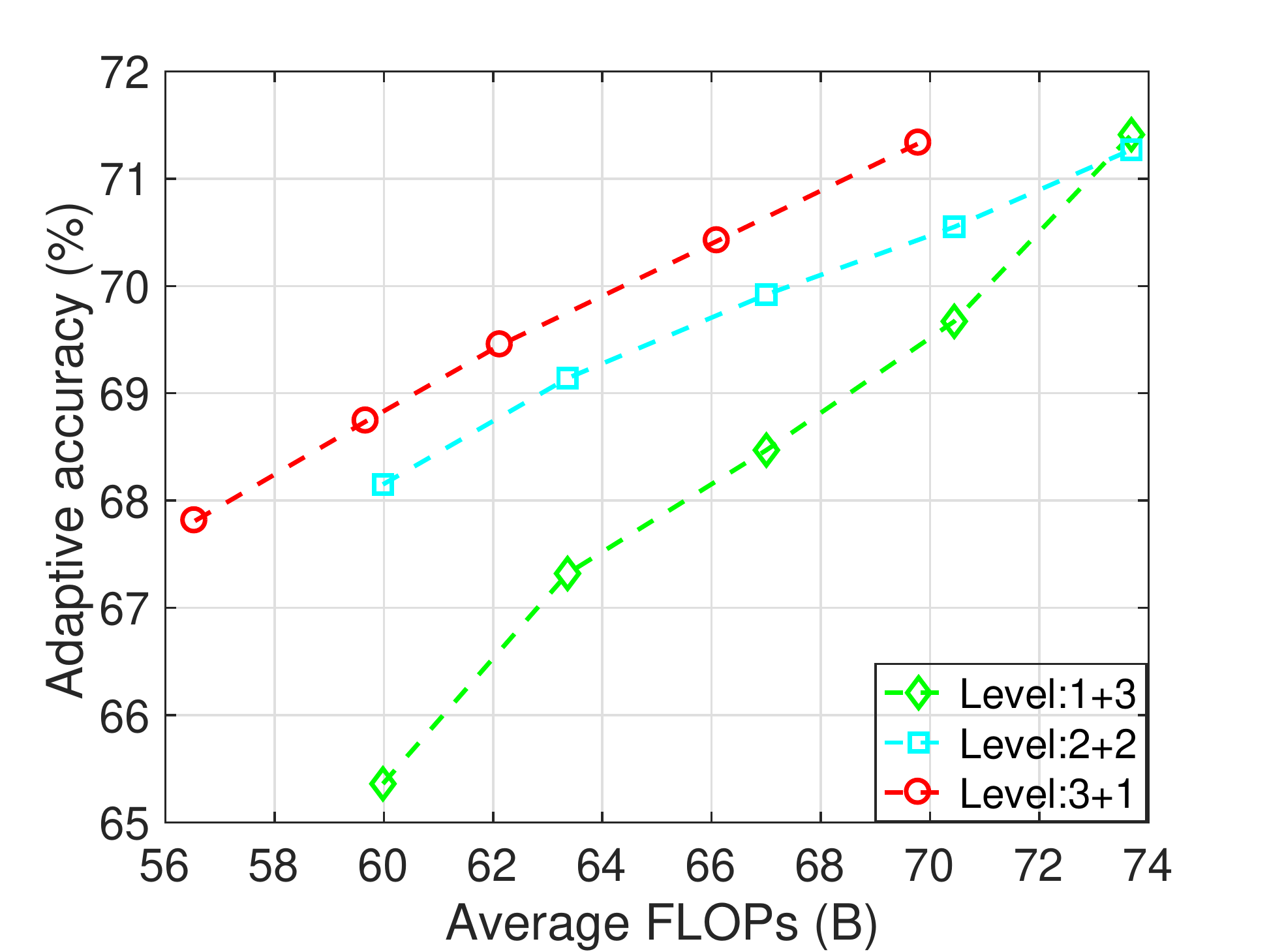}
    \label{fig_ar_a}
  }
  \subfigure[ResNet-110]{
    \includegraphics[scale = 0.21]{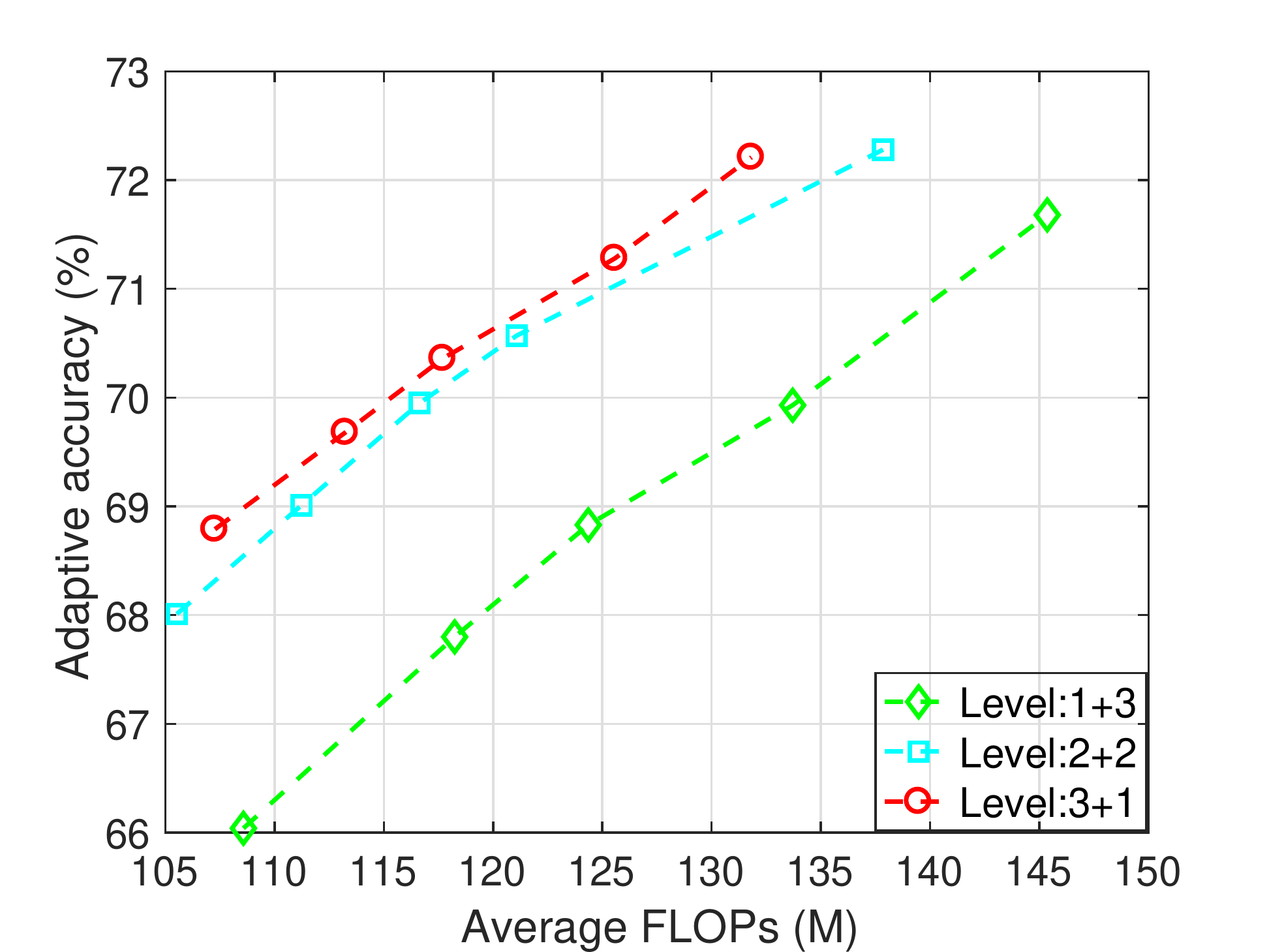}
    \label{fig_ar_b}
  }
  \subfigure[DenseNet-40-12]{
    \includegraphics[scale = 0.21]{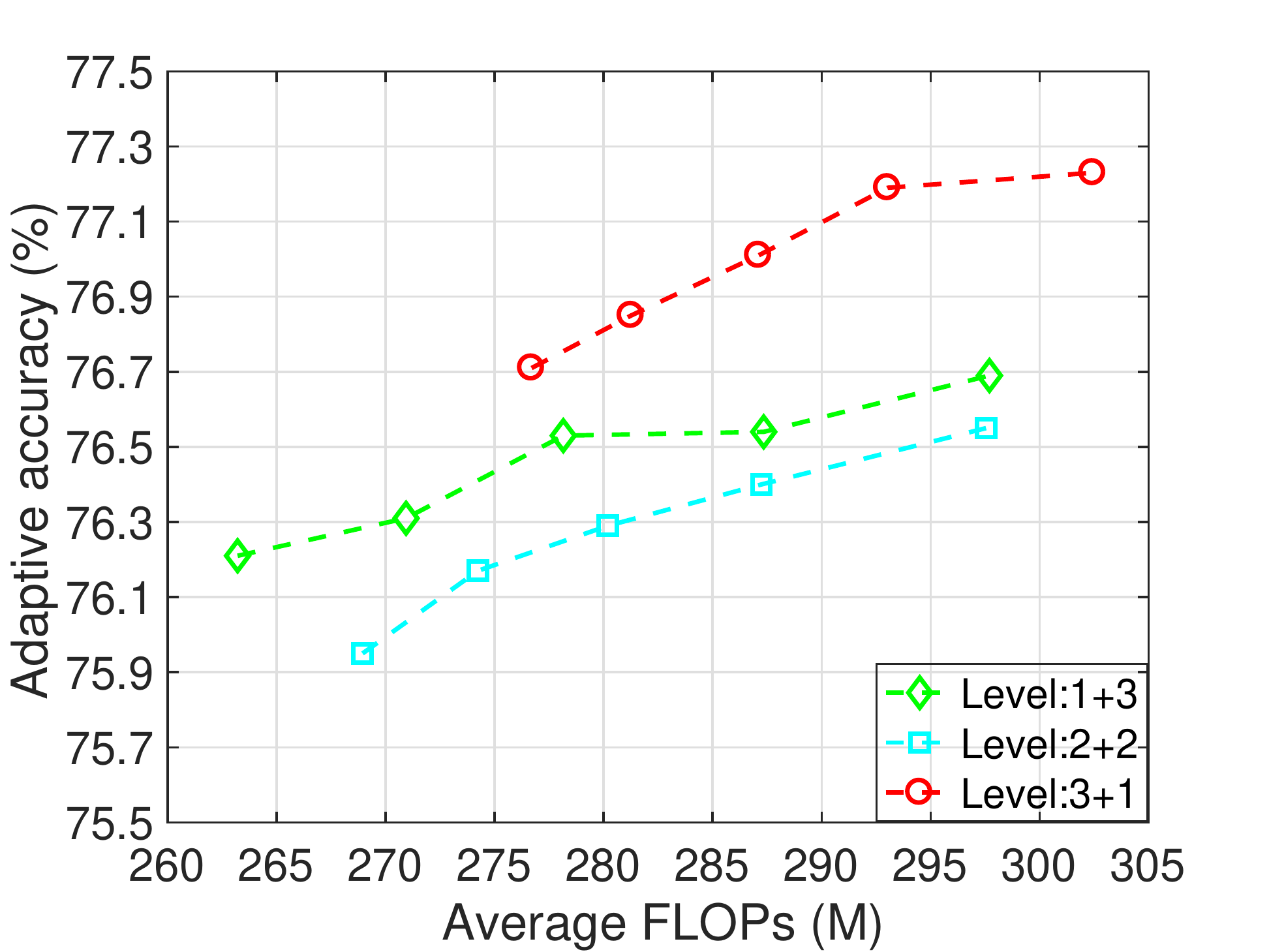}
    \label{fig_ar_c}
  }
  \subfigure[GoogleNet]{
    \includegraphics[scale = 0.21]{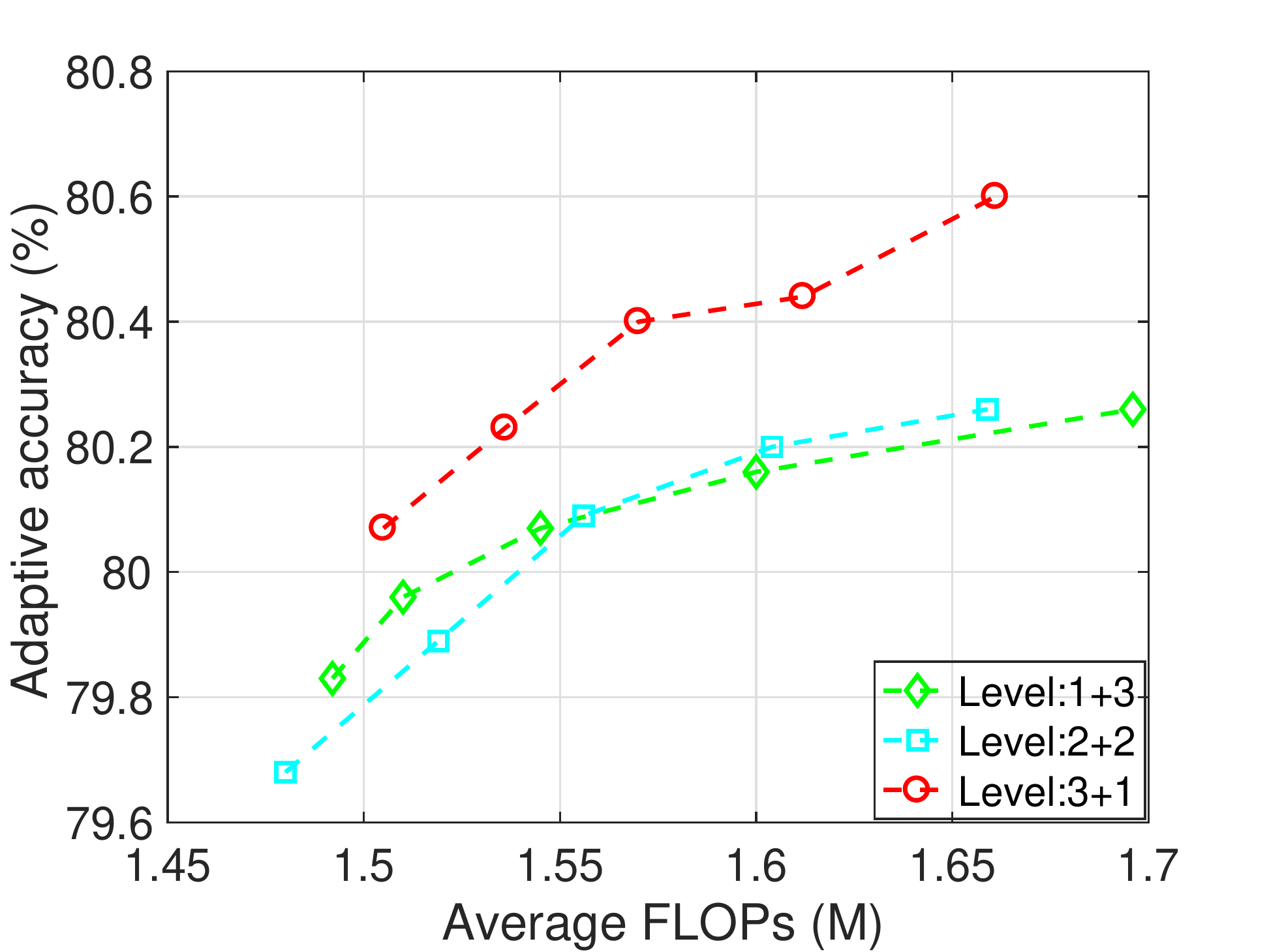}
    \label{fig_ar_d}
  }
\caption{Adaptive results of branching patterns on CIFAR-100. The complexity-decreasing branching of 3+1 achieves the best performance, compared to that on the complexity constant branching of 2+2 and the complexity-increasing branching of 1+3. 
}
\label{fig_ar}
\end{figure*}

\subsubsection{Individual Branch Classifiers} 
As shown in Table \ref{tab_resnet18_vgg},
we first look only at the branches on ImageNet and find that accuracy is directly proportional to complexity. More SE-B blocks result in more accurate branch classifiers. 
%
We also observe, as expected, that branches at higher stages are more accurate. Curiously, for complexity-increasing branches (1+2+3, 2+3+4), $f_3$'s accuracy may even exceeds $f_4$.  This likely speaks to the strength of the SE-B blocks.  Using the most blocks at the penultimate stage concentrates discriminative processing upon already high-level features, and is therefore comparable or even better than the backbone's end-stage classifier. 

Table~\ref{tab_resnet56_110} on CIFAR-100 shows that complexity-decreasing branches (3+1) achieves the highest accuracies on the first and second branches compared to complexity-increasing and constant branches.  Note that for CIFAR-100, we experiment with only two branch classifiers, \ie the branch classifier $M=3$ remains consistent with the original backbone's classifier. 
%
The exception is DenseNet-40-12, where we observe that the accuracy of $f_1$ significantly surpasses $f_3$ on DenseNet-40-12, likely due to the effectiveness of our classifier design with the feature-reuse mechanism of DenseNet. Moreover, the complexity-decreasing branch of 3+1 at the first exit already achieves performance gains over the DenseNet-40-12 baseline. 

\subsubsection{Percentage of samples exited}
We further explore the percentage of samples exited for different branching patterns under the (approximately) same adaptive accuracy. To do so, we find the combination of $\gamma$ values for each branch to obtain the lowest average FLOPs. 
For convenience of discussion, we fix the adaptive accuracy to 68.88\% and 73.36\% on ResNet-18 and VGG-16, whose results are shown in Fig.~\ref{fig_sample_exit}. 


As shown in Figs.~\ref{fig_exit_a} and ~\ref{fig_exit_b}, a large number of samples are exited at $f_1$. 35-71\% of the samples are exited at $f_1$ for the ResNet-18 backbone; this significantly surpasses the samples exited at any of the subsequent branches. On the VGG-16 backbone, this percentage is even larger, \ie 50-85\%. 
Fig.~\ref{fig_exit_c} and Fig.~\ref{fig_exit_d} shows the corresponding average FLOPS for the branching combinations in Figs.~\ref{fig_exit_a} and ~\ref{fig_exit_b} respectively.  Fig.~\ref{fig_exit_c} shows that it pays off computationally to use more complex classifiers at $f_1$.  With 6 total SE-B blocks, the complexity-decreasing branching pattern of 3+2+1 achieves the lowest average FLOPs of 1.29B, compared to 1.34B using 2+2+2 and 1.42B using 1+2+3. Additionally, using 4+3+2 is preferable to 3+2+1;
even though it has more total FLOPs, the average is still lower of 1.20B at the same adaptive accuracy. This is due to an even larger proportion of samples exited at $f_1$. Fig.~\ref{fig_exit_d} shows similar trends for VGG-16. 

Looking at the threshold $\gamma$ (see Figs.~\ref{fig_exit_e} and ~\ref{fig_exit_f}) we observe that 
complexity-decreasing branches at the same total FLOPs have the lowest threshold, \ie allows more samples to be exited earlier, especially at $f_1$. For example, the threshold $\gamma$ at 4+3+2 is set to the lowest values of (0.4, 0.4, 0.45) at $f_1$ to $f_3$, compared to (0.45, 0.45, 0.45) at 3+3+3 and (0.5, 0.45, 0.5) 
under the relative same adaptive accuracy.

\subsubsection{Adaptive Results}
Fig.~\ref{fig_af_ac} plots the adaptive accuracy on ImageNet for some fixed average FLOPs.  Note that the average FLOPS is controlled by setting different $\gamma$ at each branch.  Firstly, we observe that the larger the number of total SE-B blocks, the better adaptive results. For example, for the same average FLOPs, a branching pattern of 2+3+4 achieves the highest adaptive accuracy than a 1+2+3 branching. This trend is consistent also for constant and complexity-decreasing branching.  Overall, complexity-decreasing branching achieves the best adaptive results at the same total FLOPs, compared to complexity-increasing and constant complexity branches. For example, on VGG-16, 4+3+2 achieves the lowest average FLOPs of 7.02B (\emph{vs.} 7.66B in 3+3+3 and 8.18B in 2+3+4) at the accuracy level of 73.36\%. This is possibly due to complexity-decreasing branching having less disruptions to the feature abstraction hierarchy in the backbone. In the following section, we also use this result or combination level as the basis of our remaining analysis for Sections \ref{wed} and~\ref{branch_impact}. 
%

We also experiment the adaptive results on CIFAR-100, as shown in Fig.~\ref{fig_ar}. We find that the complexity-decreasing branching of 3+1 achieves the best performance on ResNet-56/110, DenseNet-40-12 and GoogleNet, compared to that on the complexity constant branching of 2+2 and the complexity-increasing branching of 1+3. 

\subsubsection{The Effect of Squeeze-Excitation (SE) Module on Branches}
The squeeze-excitation (SE) modules provide a form of self-attention to the branch classifiers.  We perform an ablation study without such a module in our basic SE-B block, \ie using only the bottleneck component and compare the differences on CIFAR-100 in Table~\ref{tab_se}. 
Adding the SE module significantly improves the accuracy of intermediate branches for all branching patterns, with a negligible FLOP increase (less than 0.01M FLOPs) in each branch. Overall, SE-module helps to improve the adaptive accuracy and also reduce the average FLOPs.

\begin{table}[t]
\scriptsize
\caption{Comparisons W \& W/O SE-module of branches on CIFAR-100.}
\begin{center}
\scalebox{0.75}{
\bgroup
\def\arraystretch{1.15}
\begin{tabular}{|c|c|c|c|c|c|c|c|c|c|}
\hline
\multirow{2}*{Model} & \multirow{2}*{SE} & \multicolumn{3}{c|}{ Branch Classifier Accuracy(\%) } & \multicolumn{3}{c|}{ Branch Classifier FLOPS(M) } & \multicolumn{2}{c|}{Adaptive}\\
\cline{3-10}
 &  & $f_1$ & $f_2$ & $f_3$ & $f_1$ & $f_2$ & Total & Acc\% & AF(M) \\
\hline
1+3 & Yes & 59.87 & 69.20 & 73.02 & 45.89 & 91.51 & \multirow{6}*{132.80} & 70.57 & 73.58 \\
ResNet-56 & No & 54.45 & 63.83 & 71.84 & 45.89 & 91.50 & & 68.42 & 82.52 \\
\cline{1-7}
\cline{9-10}
2+2 & Yes &  64.46 & 69.26 & 72.16 & 46.43 & 91.51 & & 70.10 & 65.69 \\
ResNet-56 & No & 55.20 & 64.47 & 72.58 & 46.43 & 91.50 & & 68.39 & 81.01  \\
\cline{1-7}
\cline{9-10}
3+1 & Yes & 64.52 & 69.32 & 72.80 & 46.97 & 91.51 & & 70.70 & 67.07 \\
ResNet-56 & No & 57.84 & 66.28 & 72.47 & 46.97 & 91.50 & & 68.08 & 77.00  \\
\hline
\hline
1+3 & Yes & 61.53 & 70.03 & 73.65 & 88.36 & 176.44 & \multirow{6}*{260.20} & 70.74 & 138.03  \\
ResNet-110 & No & 56.60 & 68.08 & 74.33 & 88.36 & 176.44 & & 69.71 & 147.10  \\
\cline{1-7}
\cline{9-10} 
2+2 & Yes & 64.96 & 70.12 & 73.70 & 88.90 & 176.44 & & 71.22 & 124.66 \\
ResNet-110 & No & 59.07 & 67.04 & 74.16 & 88.90 & 176.44 & & 69.46 & 143.75 \\
\cline{1-7}
\cline{9-10}
3+1 & Yes & 65.14 & 70.81 & 74.22 & 89.44 & 176.44 & & 71.53 & 127.20 \\
ResNet-110 & No & 56.87 & 65.72 & 73.49 & 89.44 & 176.44 & & 68.86 & 148.88 \\
\hline
\end{tabular}
\egroup
}
\end{center}
\label{tab_se}
\end{table}

\begin{figure*}[t]
\centering
  \subfigure[MSDNet on CIFAR-100]{
    \includegraphics[scale = 0.21]{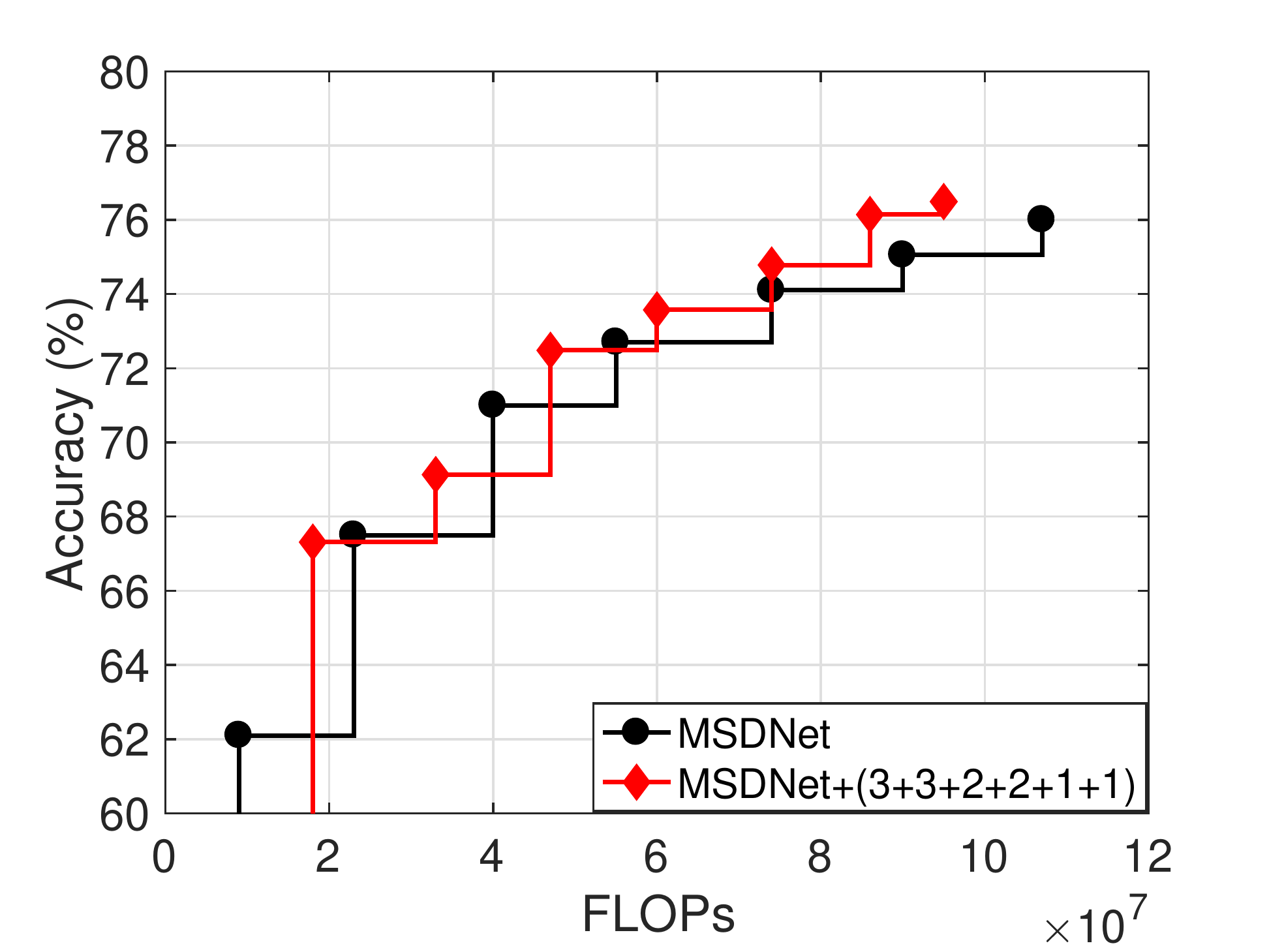}
    \label{fig_r_a}
  }
  \subfigure[ResNet-56 w \& w/o SE on CIFAR-100]{
    \includegraphics[scale = 0.21]{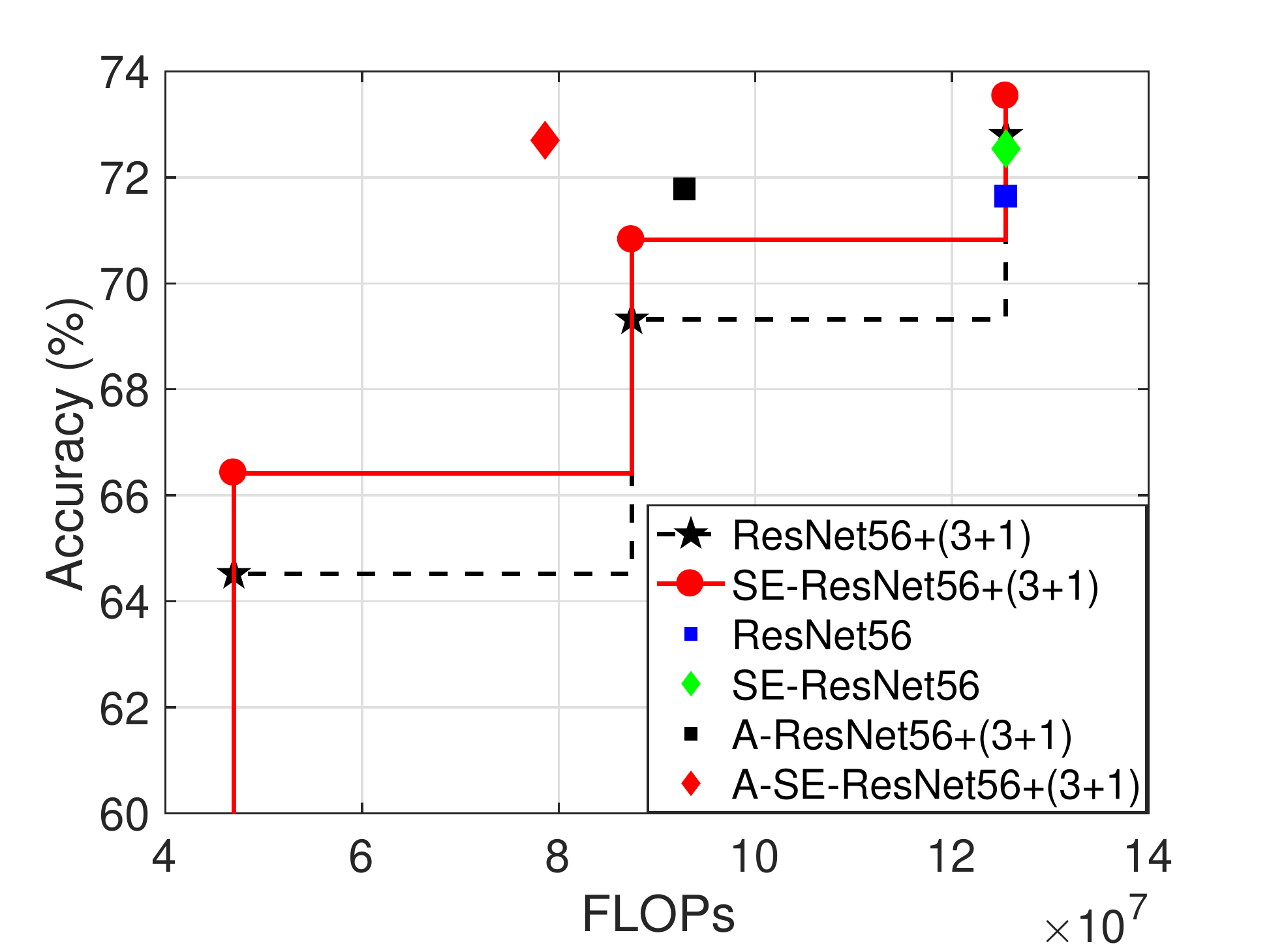}
    \label{fig_r_b}
  }
  \subfigure[MSDNet on ImageNet]{
    \includegraphics[scale = 0.21]{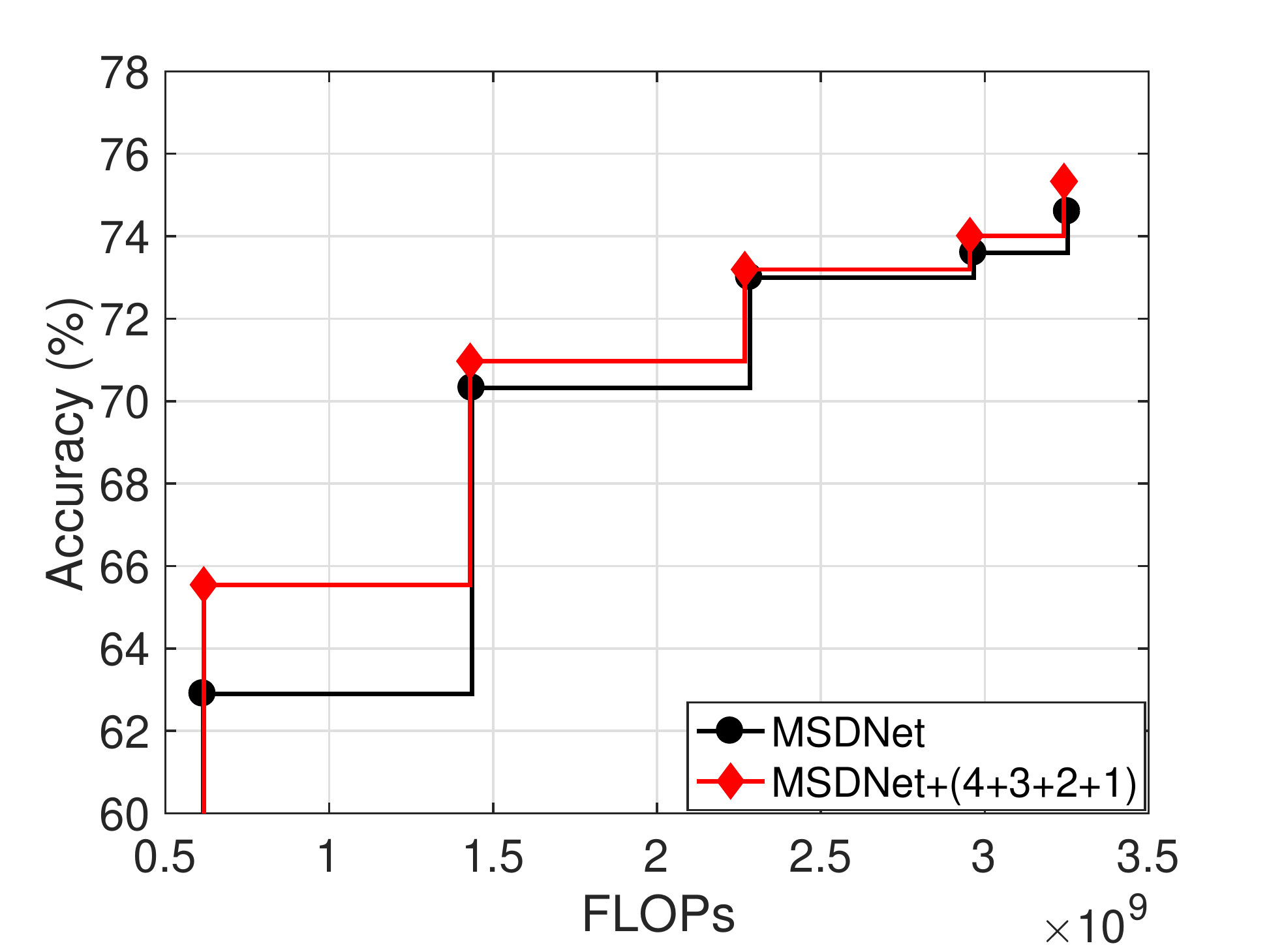}
    \label{fig_r_c}
  }
  \subfigure[MobileNetV2 on ImageNet]{
    \includegraphics[scale = 0.21]{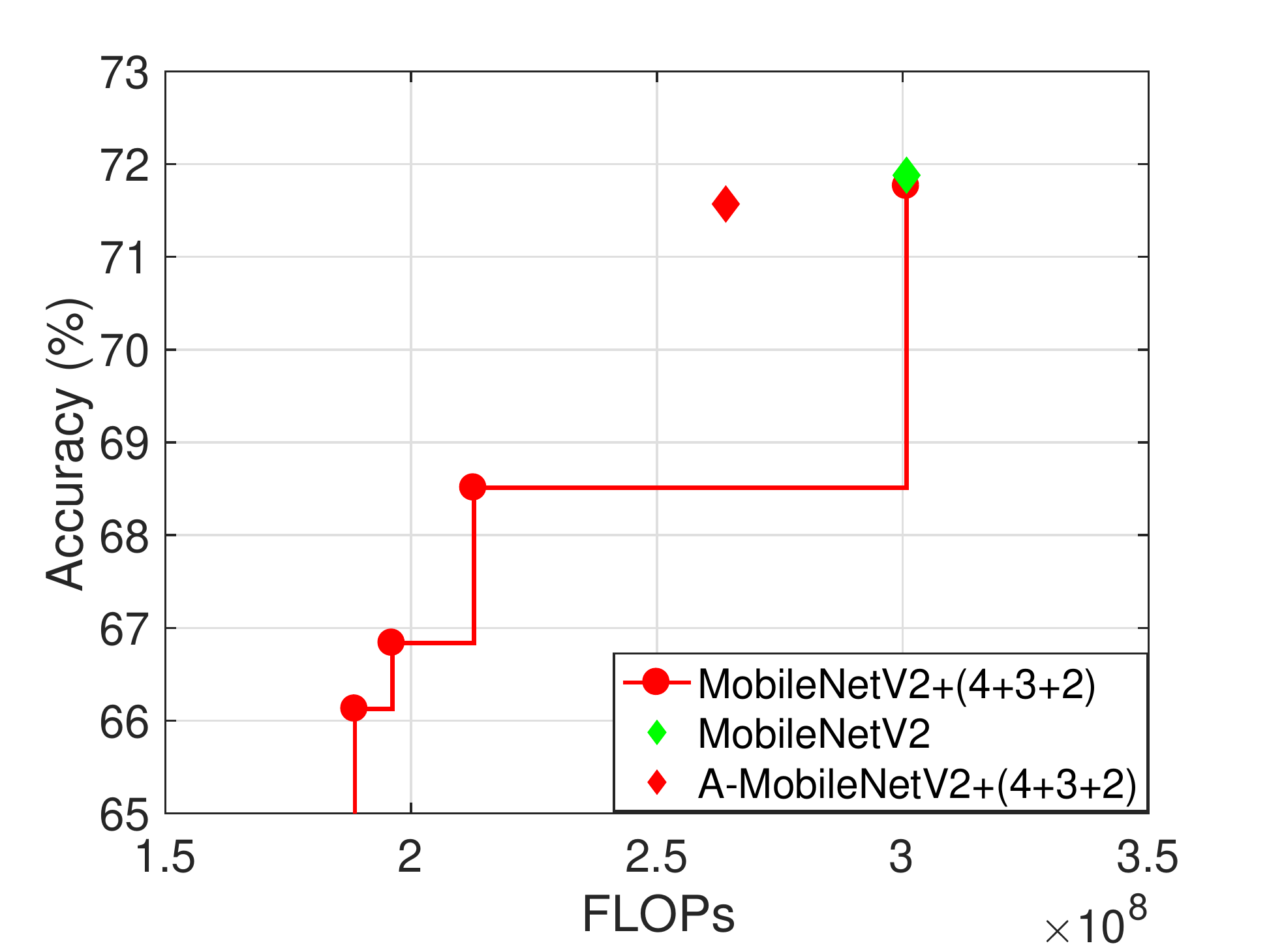}
    \label{fig_r_d}
  }
\caption{Our branching on different backbones on CIFAR-100 and ImageNet. X+(Y) denotes backbone X with branching (Y); A- indicates the adaptive result.
}
\label{fig_r}
\end{figure*}

\subsection{Training and Knowledge Distillation}
\label{wed}

We compare the effect of different training strategies in Table~\ref{tab_distillation} and add two naive strategies to give some insight.  One is a \emph{branches-only} training (BO), where $\{\phi_M, \theta_M\}$ for $\text{CE}_M$ is learned first and frozen; only $\theta_i$ with $\sum_i^{M-1}\text{CE}_i$ are learned after.  A second variant is \emph{stage-wise} (SW), in which $\{\phi_i,\theta_i\}$ are learned 
with $\text{CE}_i$ for $i$ sequentially from 1 to $M$, while keeping shared weights $\phi_i \subset \phi_{i+1}$ fixed.  

We find that the cooperative training (Coop) by minimizing Eq. \ref{eq:loss} where all $\{\phi_i, \theta_i\}$ are learned jointly achieves the best performance-cost trade-off. This result is consistent to tree learner~\cite{kumar2017resource} for learning all nodes jointly.  The naive baselines show that it is important to balance the learning so that the backbone is useful for all branches. In \emph{branches only}, the backbone is optimized for $f_4$, hence the lower performance for earlier classifiers $f_1$ and $f_2$ and the higher adaptive accuracy cost. In \emph{stage-wise} training, the backbone parameters learned greedily for each stage, so the incompatibilities compound, hence the much lower $f_4$ and adaptive accuracies.

We also compare in Table \ref{tab_distillation} 
our weighted ensemble distillation (WED) with other distillation methods that use different teachers $y_t$:
dark knowledge (DK)~\cite{hinton2015distilling} uses a pre-trained baseline, OFA uses the end-stage classifier \cite{li2019improved} and mean ensemble (MED)~\cite{phuong2019distillation} averages all branches.  In general, knowledge distillation improves branch and adaptive accuracies while decreasing the adaptive FLOPs. However, improvements are directly tied to the accuracy of the teachers.  In the case of GoogLeNet, DK~\cite{hinton2015distilling} and OFA~\cite{li2019improved} drops even below the standard cooperative training, due to their low teacher accuracies.  In contrast, the ensemble approaches MED~\cite{phuong2019distillation} and our WED yield strong teachers. WED's $y_t$ has an accuracy of over 81\%, which are then transferred over to the branches.

\begin{table}[t]
\footnotesize
\caption{Comparison of training strategies. BO and SW are the naive branches-only and stage-wise training respectively. Coop is standard cooperative training. DK~\cite{hinton2015distilling}, OFA~\cite{li2019improved} and MED~\cite{phuong2019distillation} are competing knowledge distillation methods. WED is ours (see Eq.~\ref{eq_kd}) and $y_t$ denotes teacher accuracy.  
} 
\begin{center}
\scalebox{0.83}{
\bgroup
\def\arraystretch{1.15}
\begin{tabular}{|c|c|c|c|c|c|c|c|c|}
\hline
\multirow{2}*{Model} & \multirow{2}*{Method} & \multicolumn{5}{c|}{ Classifiers' Accuracy (\%) } & \multicolumn{2}{c|}{Adaptive}\\
\cline{3-9}
 &  & $f_1$ & $f_2$ & $f_3$ & $f_4$ & $y_t$ & Acc & A-F \\
\hline
\multirow{7}{*}{\begin{tabular}[c]{@{}c@{}}  4+3+2\\  (ResNet-18\\ on ImageNet)\end{tabular}} & BO & 64.3& 66.2 & 67.9& 69.8& - & 69.2\% & 1.3B \\ 
 & SW & 65.5 & 66.6 & 67.8 & 67.6 & - & 68.3\% & 1.3B \\
& Coop & 65.4 & 67.1 & 69.2 & 69.3 & - & 69.3\% & 1.3B \\
 & DK & 65.4 & 67.1 & 69.5 & 69.2 & 69.8 & 69.2\% & 1.3B \\
 & OFA & 66.4 & 67.9 & 70.0 & 70.0 & 70.0 & 70.3\% & 1.3B\\
 & MED & 66.4 & 67.9 & 70.0& 69.9& 71.8 & 70.2\% & 1.3B \\ 
 & WED & 66.5 & 67.9 & 70.1 & 70.0 & 72.1 & \textbf{70.4\%} & 1.3B \\
\hline
\multirow{5}{*}{\begin{tabular}[c]{@{}c@{}}3+1\\ (GoogLeNet\\ on CIFAR-100)\end{tabular}}  
& Coop & 77.3 & 80.0 & 79.0 & - & -  & 78.1\% & 1.4B  \\ 
 & DK & 76.8 & 79.9 & 78.1 & - & 78.4 & 78.0\% & 1.4B \\
 & OFA & 76.5 & 79.8 & 78.3 & - & 78.3 & 77.6\% & 1.4B \\
 & MED & 76.6& 79.5 & 78.6 & - & 81.1 & 77.8\% & 1.4B \\
 & WED & 76.9 & 79.7 & 78.9 & -  & 81.1 & \textbf{78.2\%}  & 1.4B  \\
\hline
\end{tabular}
\egroup
}
\label{tab_distillation}
\end{center}
\end{table}

\begin{figure*}[t]
\centering
  \subfigure[]{
    \includegraphics[scale = 0.255]{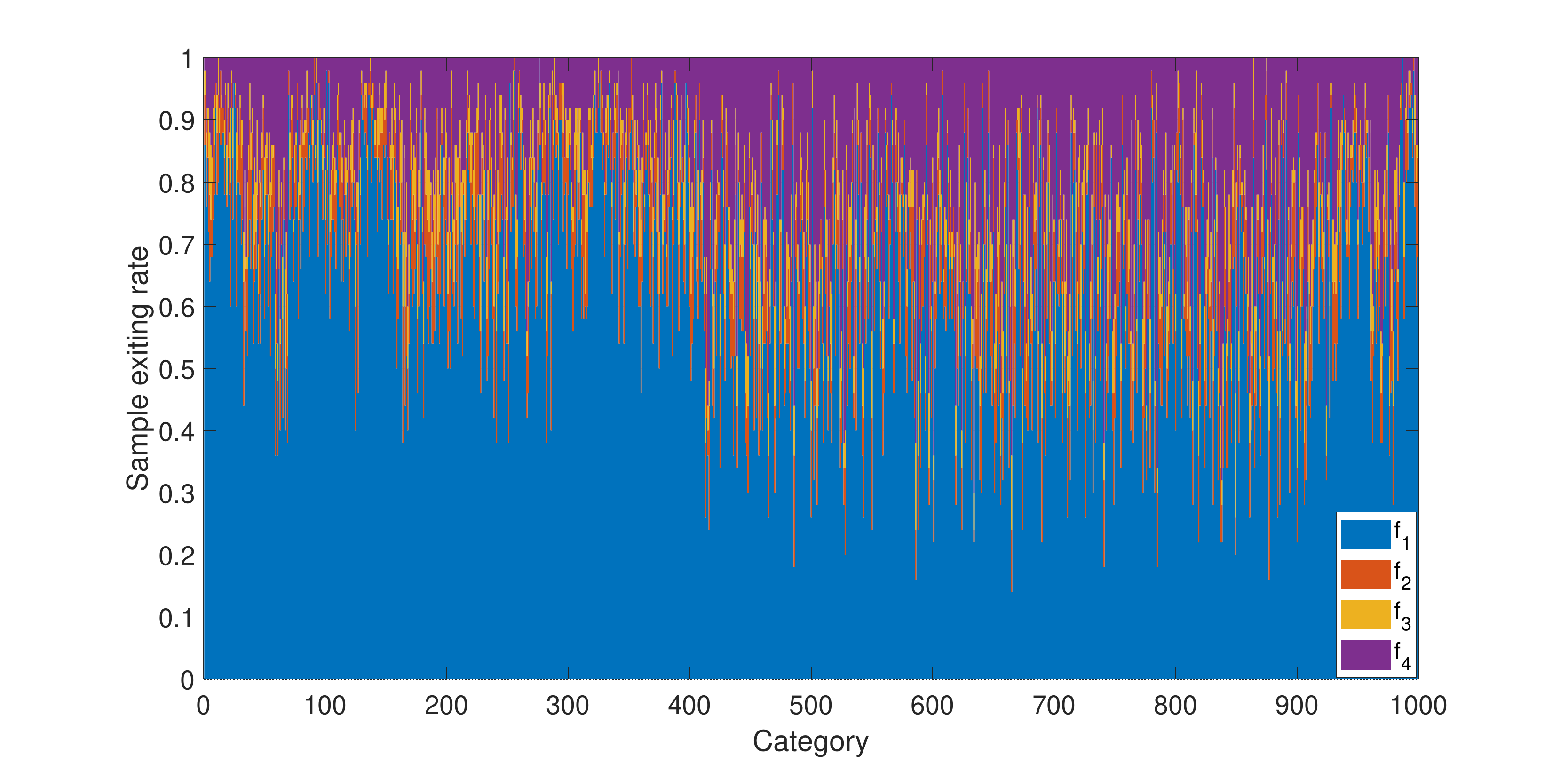}
    \label{fig_v_se}
  }
  \subfigure[]{
    \includegraphics[scale = 0.255]{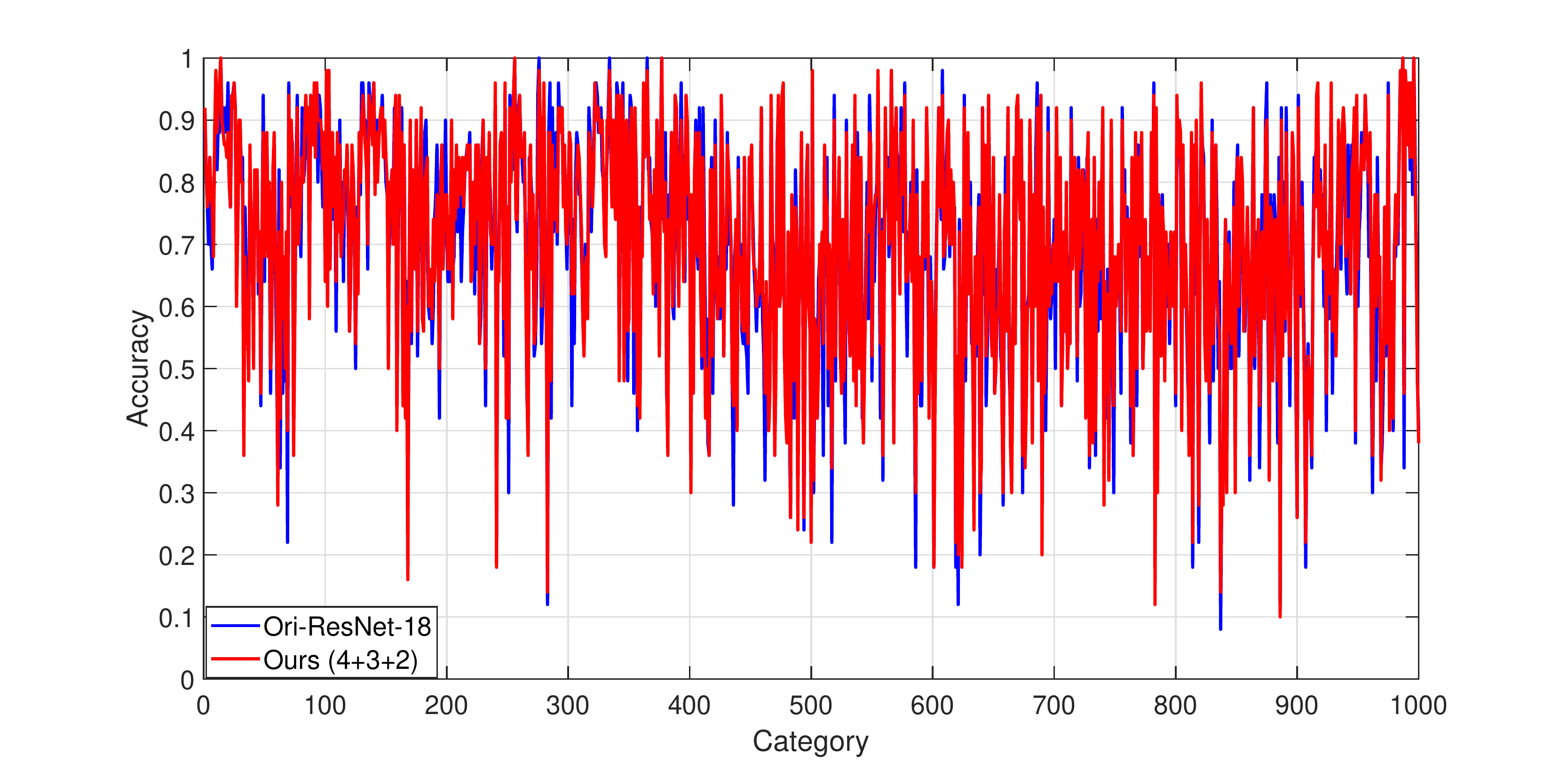}
    \label{fig_v_ac}
  }
\caption{The relationship between accuracy and sample exiting rate at each branch on ResNet-18 using ours (4+3+2). (a) Sample exiting rate of each branch at 1000 classes. (b) Accuracy of each class on original ResNet-18 and our branch (4+3+2). }
\label{fig_v1}
\end{figure*}

\subsection{Comparison with Other Branching Methods}
We compare with two recent branching architectures from the literature: SCAN~\cite{zhang2019scan} and  MSDNet~\cite{huang2017multi}. SCAN constructs scalable networks with highly elaborate branch classifiers that use one large convolution-deconvolution attention module followed by bottleneck structure and one convolution with a linear projection.  MSDNet features a customized multi-scale backbone with simple branch classifiers that use two convolutions and a linear projection. 
For fair comparison, we keep backbones the same while comparing our branch classifiers to the others.    

\textbf{CIFAR-100.} We compare with MSDNet with their own proposed backbone as well as ResNet-56 and SE-ResNet-56 with SE-blocks.  On the MSDNet backbone, we apply 6 of our branches (3+3+2+2+1+1) placed in the same locations as the original MSDNet\footnote{https://github.com/kalviny/MSDNet-PyTorch}. On the ResNet backbones, we apply a (3+1) branching.  Results from Fig.~\ref{fig_r_a} and ~\ref{fig_r_b} show that relative gains from our branch classiﬁers on the same backbones are always present. The MSDNet backbone and SE-ResNet-56 are stronger backbones than ResNet-56 and shift the overall performance upwards. For SE-ResNet-56, adding branching improves the adaptive accuracy by 0.16\% (\emph{vs.} SE-ResNet-56) while reducing about 47M FLOPs on average.  
Compared to MSDNet, we use slightly more FLOPS in the earlier branches, but our accuracy is signiﬁcantly higher than MSDNet’s reported results (see Fig. \ref{fig_r_b}).

\textbf{ImageNet.} We first compare the branches of MSDNet and branches of SCAN on the same ResNet-18/34/50 backbones and 
show results in Table \ref{tab_resnet18_34}. It appears incomplete as the original SCAN did not report the ResNet-34 result. We report additionally a baseline with no branching.  On all three ResNet backbones, our 4+3+2 branching, 
when trained with weighted ensemble distillation, achieves the highest accuracy for all the branch classifiers.  We also report the highest adaptive accuracy and the lowest average FLOPs.  
Compared to the baseline, we use only 71.3\%, 54.6\% and 72.4\% of the FLOPs but achieve performance gains of 0.51\%, 0.19\% and 0.22\% improvement over the original ResNet-18, ResNet-34 and ResNet-50 respectively. Compared to the branches used by MSDNet and SCAN, our method achieves the best performance both in each classifier's accuracy, adaptive accuracy and average FLOPs.

We further investigate by appending our branches onto MSDNet and MobileNetV2 backbones.  For the MSDNet backbone, in the same (four) locations as the original MSDNet.  We use a (4+3+2+1) branching and as shown in Fig. \ref{fig_r_c}, adding our branches to the MSDNet backbone has improved performance with almost negligible increase in FLOPS.  The most pronounced difference is on the first exit, which is expected, as we place the deepest branch there. Fig. \ref{fig_r_d} shows that on the compact backbone MobileNetV2, we achieve a reduction of 37M average FLOPs with a negligible accuracy drop, compared to the original MobilenetV2. In particular, the first branch classifier allows 55\% samples exiting with 88.86\% accuracy (only consider these 55\% samples), which significantly higher than the original MobileNetV2 of 71.88\%.


\begin{figure}[t]
\centering
  \subfigure{
    \includegraphics[scale = 0.27]{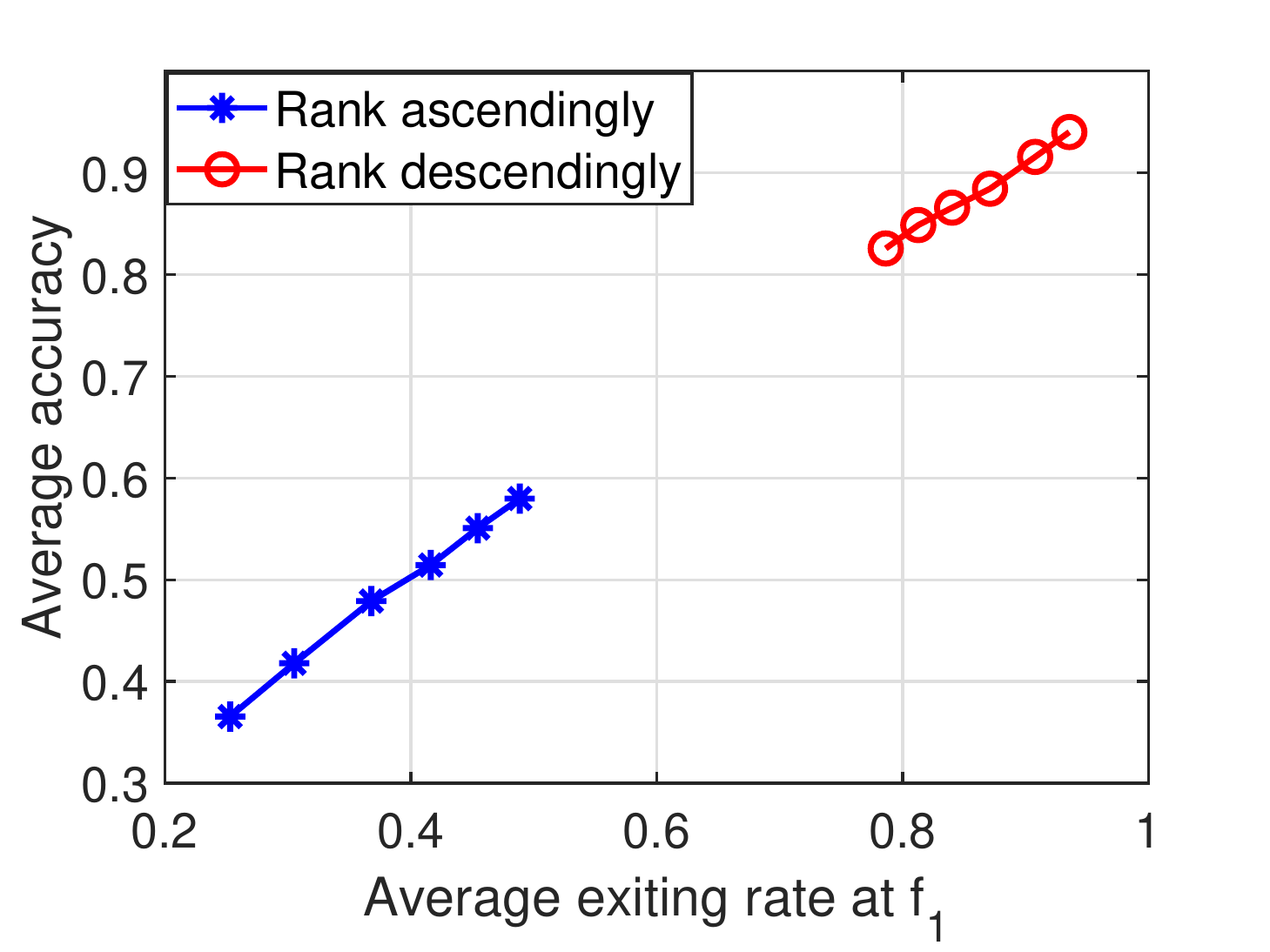}
    \label{fig_f1_acc1}
  }
  \subfigure{
    \includegraphics[scale = 0.27]{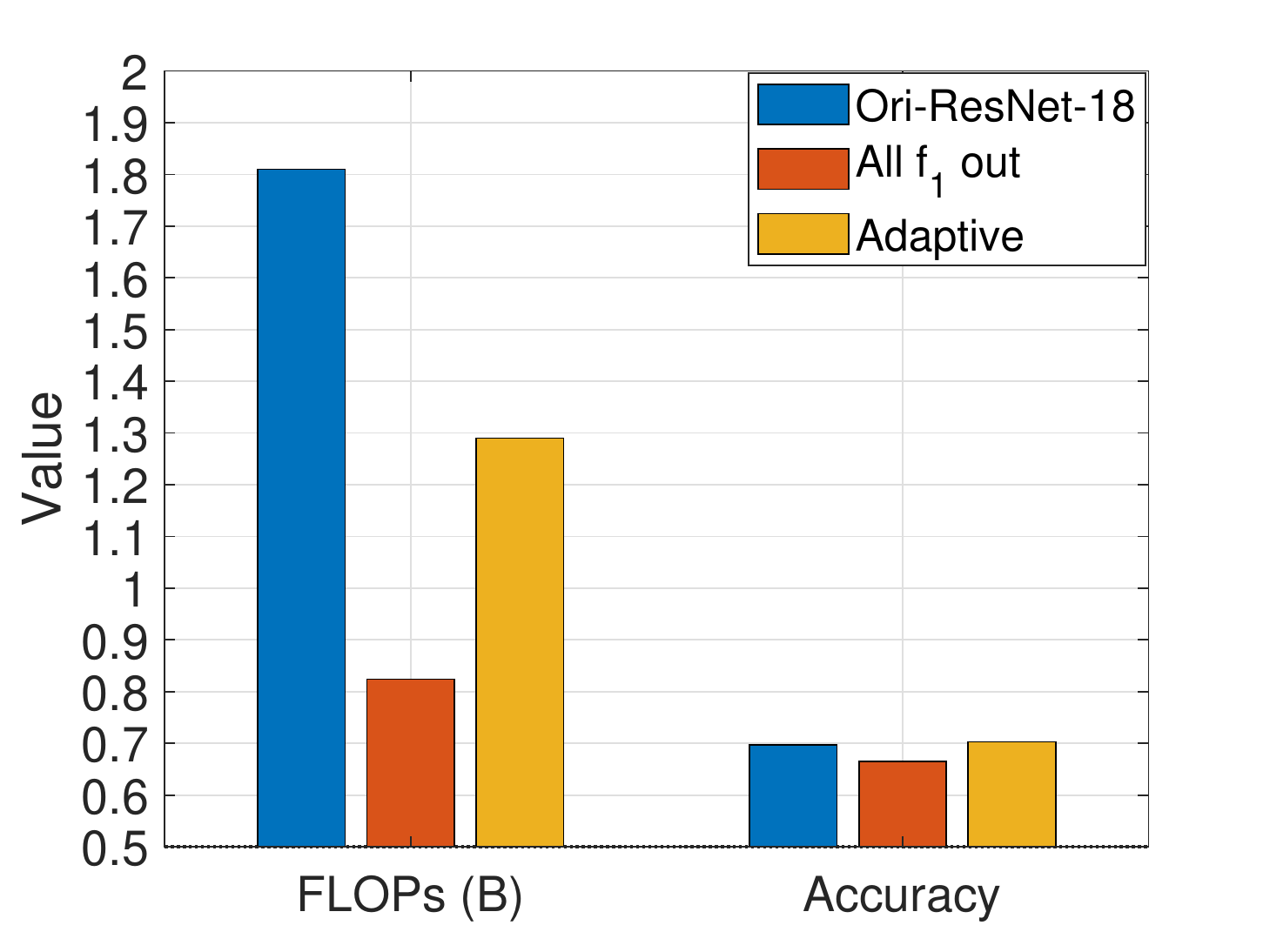}
    \label{fig_f1_acc2}
  }
\caption{The effect of $f_1$ exiting rates with respect to the accuracy on ResNet-18. Left: The relationship between the average $f_1$ exiting rates and the average accuracy based on ascending and descending sorting and evaluated from a group of 50, 100, 200, 300, 400 and 500 classes. Right: The FLOPs and accuracy comparison among the original ResNet-18, all samples exiting from $f_1$ and the adaptive result, which are denoted by Ori-ResNet-18, All $f_1$ out and Adaptive, respectively.}
\label{fig_f1_acc}
\end{figure}

\begin{table}[t]
\footnotesize
\caption{Results of different branch classifiers on ImageNet. }
\begin{center}
\scalebox{0.83}{
\bgroup
\def\arraystretch{1.15}
\begin{tabular}{|c|c|c|c|c|c|c|c|}
\hline
\multirow{2}*{Backbone} & \multirow{2}*{Branching} & \multicolumn{4}{c|}{ Classifiers' Accuracy (\%) } & \multicolumn{2}{c|}{Adaptive} \\
\cline{3-8}
 &  & $f_1$ & $f_2$ & $f_3$ & $f_4$ & Acc\% & A-F(B) \\
\hline
\multirow{4}*{ResNet-18} & no branches &  - & - & - & 69.76  & - & 1.81 \\ 
& MSDNet & 41.19 & 55.21 & 64.28 & 68.38 & 66.00  & 1.36  \\
& SCAN & 48.25 & 58.00 & 65.32 & 69.32 & - & - \\
& Ours (4+3+2) & 66.55 & 67.94 & 70.05 & 70.02  & \textbf{70.27} & \textbf{1.29} \\ 
\hline
\multirow{3}*{ResNet-34} & no branches &  - & - & - & 73.30 & - &  3.66 \\ 
& MSDNet & 42.87 & 58.56 & 70.35  & 72.68 & 70.19 & 2.44 \\
& Ours(4+3+2) & 66.97 & 69.66 & 73.77 & 74.23 & \textbf{73.49} & \textbf{2.00} \\ 
\hline
\multirow{3}*{ResNet-50} & no branches &  - & - & - & 76.15 & - &  4.09 \\ 
& SCAN & 53.86 & 66.54 & 73.57 & 75.88 & - & - \\
& Ours (4+3+2) & 68.01 & 70.86 & 75.81 & 77.21 & \textbf{76.37} & \textbf{2.96} \\
\hline
\end{tabular}
\egroup
}
\label{tab_resnet18_34}
\end{center}
\end{table}


\begin{figure}[t]
\centering
    \includegraphics[scale = 0.32]{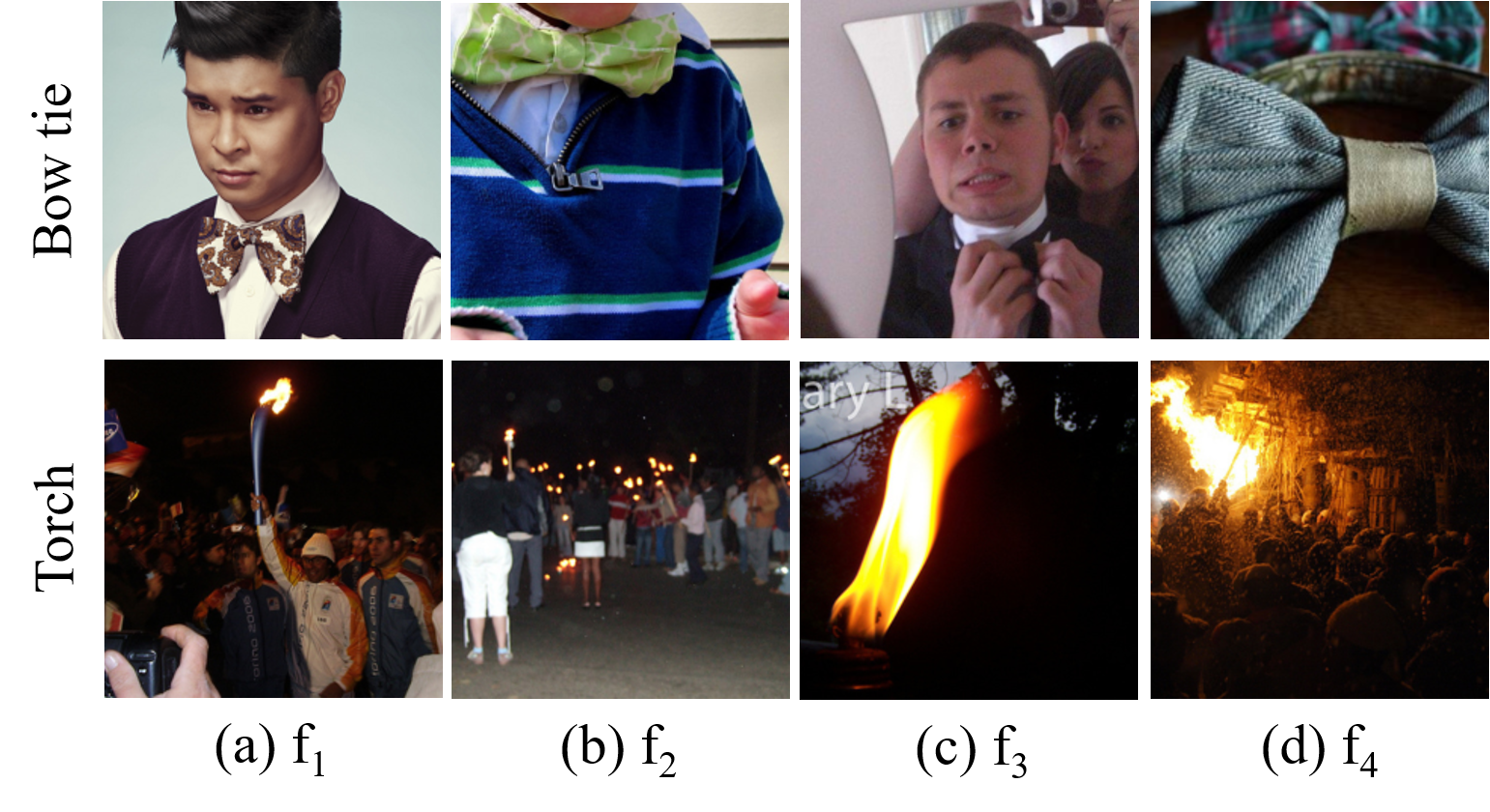}
\caption{Visualization of adaptive inference on ResNet-18. For the classes of ``bow tie'' and ``torch'', dynamic inference achieves both 12\% accuracy gains on these two classes over all samples exiting at $f_1$, \ie 72\% \emph{vs.} 60\%.}
\label{visualization}
\end{figure}

\subsection{Visualization}
We further explore the visualization of ResNet-18 at different classes using our branch of 4+3+2\footnote{The overall performance has been evaluated in Table~\ref{tab_resnet18_34}.}. As shown in Fig.~\ref{fig_v1}, test images at each class on ImageNet 2012 have different sample exiting rates at four branches, which indicates recognition difficulty exits both on inter-class and intra-class images. We also observe that sample exiting rates from $f_1$ to $f_4$ affect the adaptive accuracy at each class. We conjecture the accuracy at each layer is mainly related to the exiting rates of $f_1$.

To evaluate the above assumption, we first sorting the the exiting rates of $f_1$ from all classes in an ascending order. Then, we calculate the average exiting rates of $f_1$ and average accuracy from a group of 50, 100, 200, 300, 400 and 500 classes ranking by the front, as well as the corresponding information ranking by the bottom. As shown in Fig.~\ref{fig_f1_acc1}, we observe that the more exiting rate at $f_1$ produces higher accuracy both on the ascending and descending sorting. Although the average exiting rates of $f_1$ and average accuracy are positive correlated, it will lead to the significant whole accuracy degeneration if all samples directly exiting at $f_1$ (see in Fig.~\ref{fig_f1_acc2}). For comparison, adaptive inference achieves the best trade-off between FLOPs and accuracy. Moreover, dynamic inference are more suitable to the hierarchical understanding of images. As shown in Fig.~\ref{visualization}, the dynamic inference system can effectively recognize the easy and difficult images. The easy images can be successfully classified by the shorter inference pathway, while allocating the longer pathway to recognize the difficult ones.

\begin{figure*}[t]
\centering
    \subfigure{
    \includegraphics[scale = 0.24]{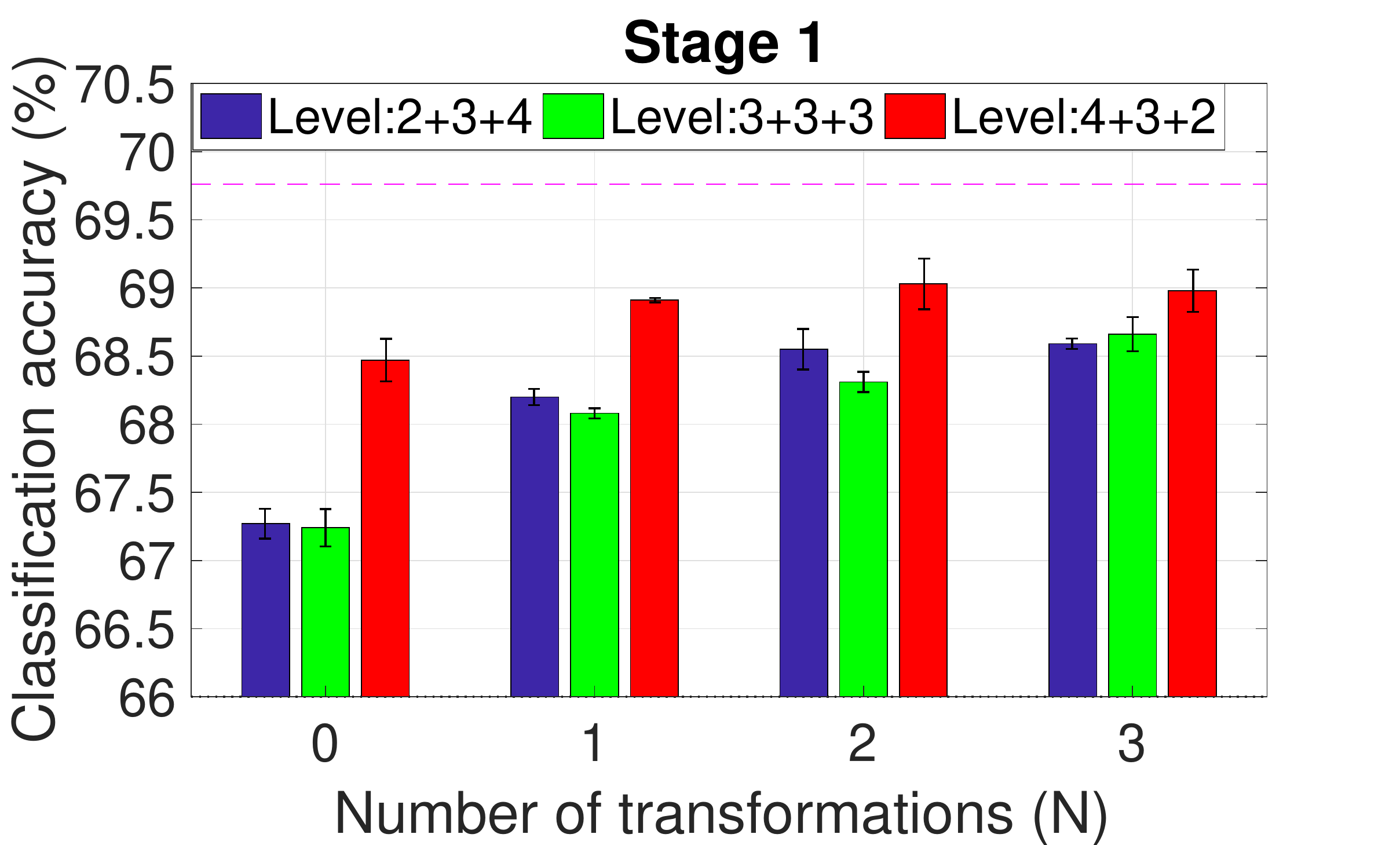}
    \includegraphics[scale = 0.24]{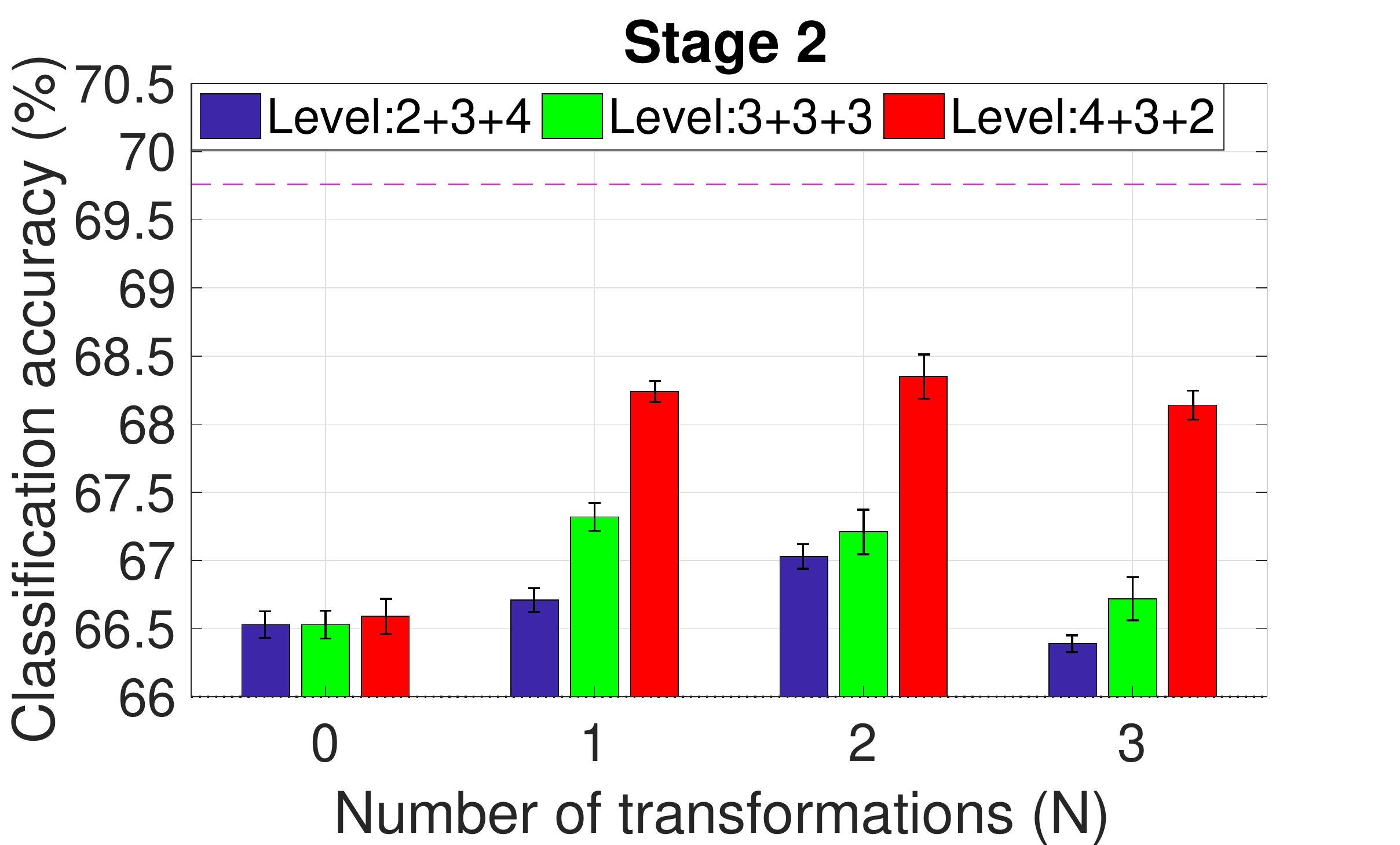}
    \includegraphics[scale = 0.24]{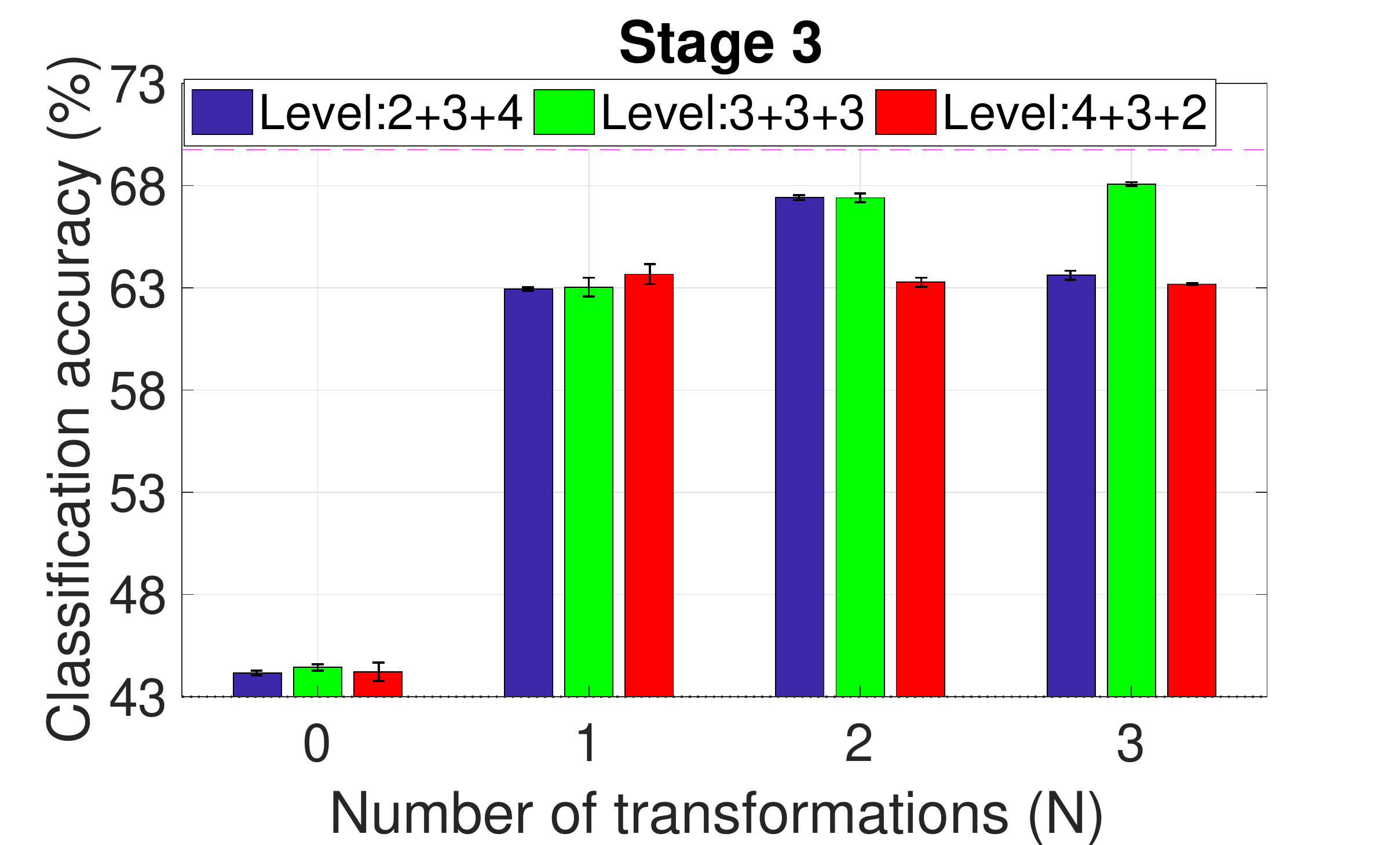}
    }
 \caption{Classification accuracy with reconstructed baseline features from different branching networks on ResNet-18. We run each reconstructed model 3 times and plot the average accuracy and standard deviation.  The magenta dashed line denotes the original baseline accuracy.
 }
 \label{fig_resnet_complexity}
\end{figure*}

\begin{figure*}[t]
\centering
    \subfigure{
    \includegraphics[scale = 0.22]{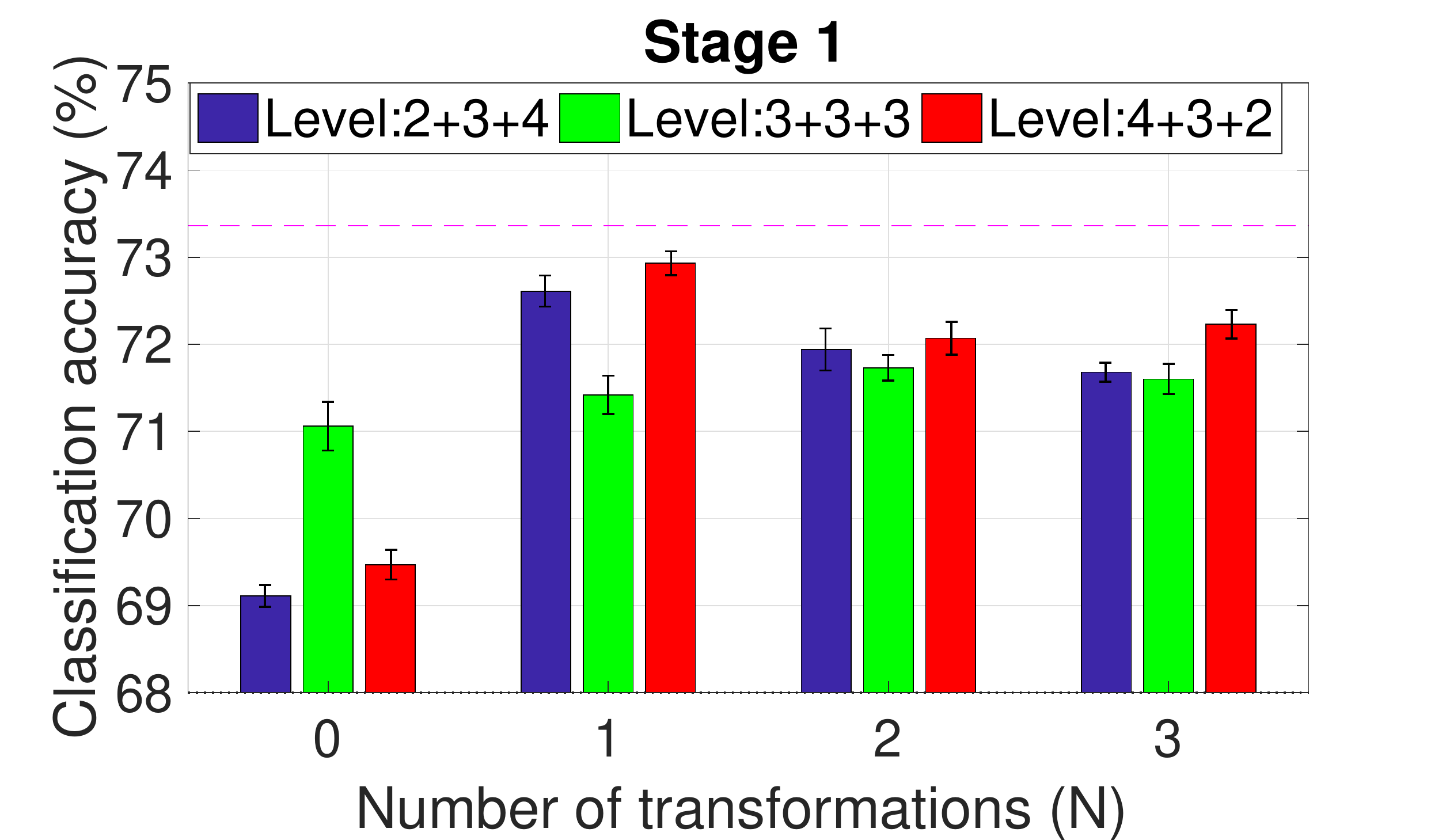}
    \includegraphics[scale = 0.22]{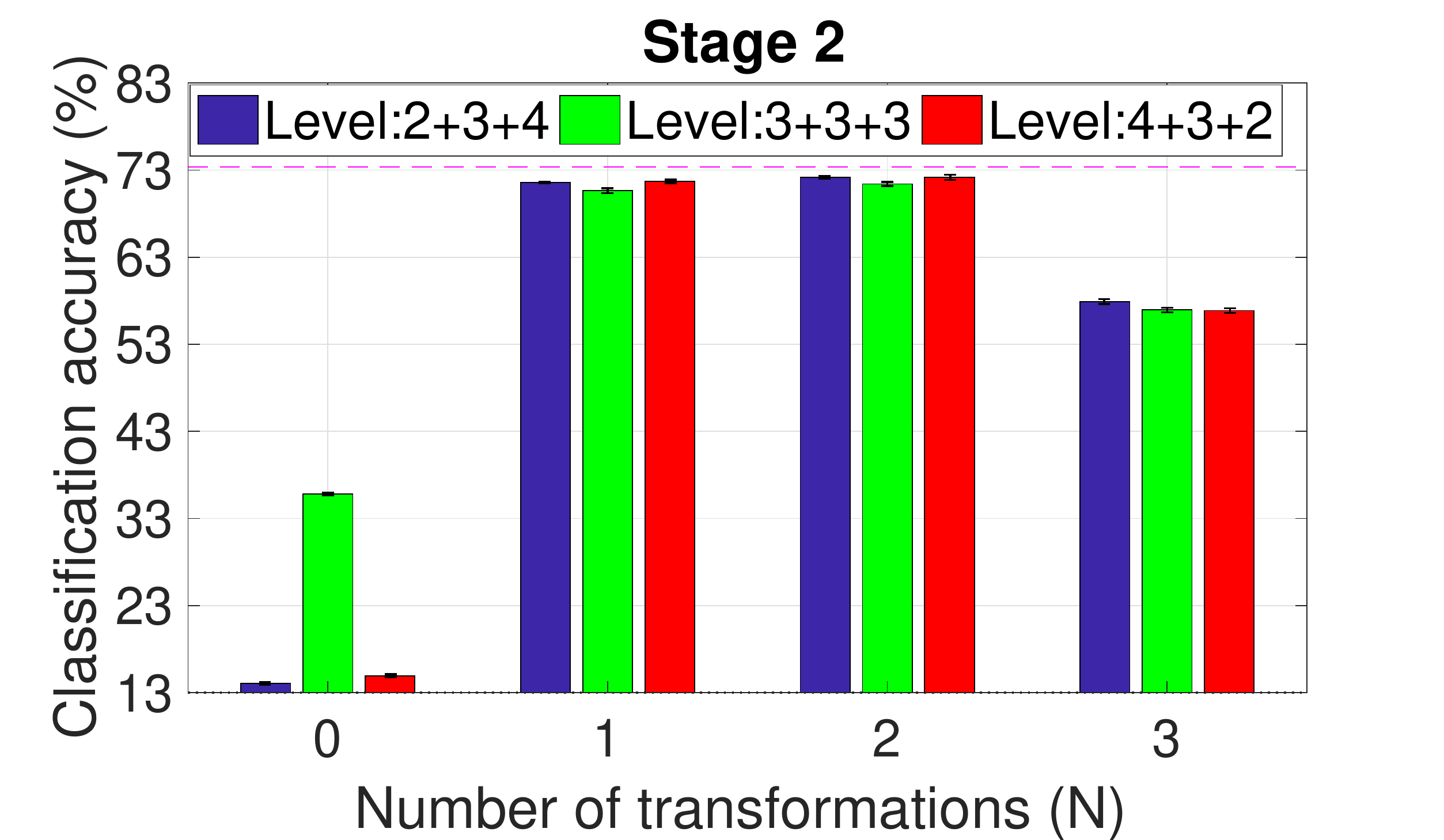}
    }
    \subfigure{
    \includegraphics[scale = 0.22]{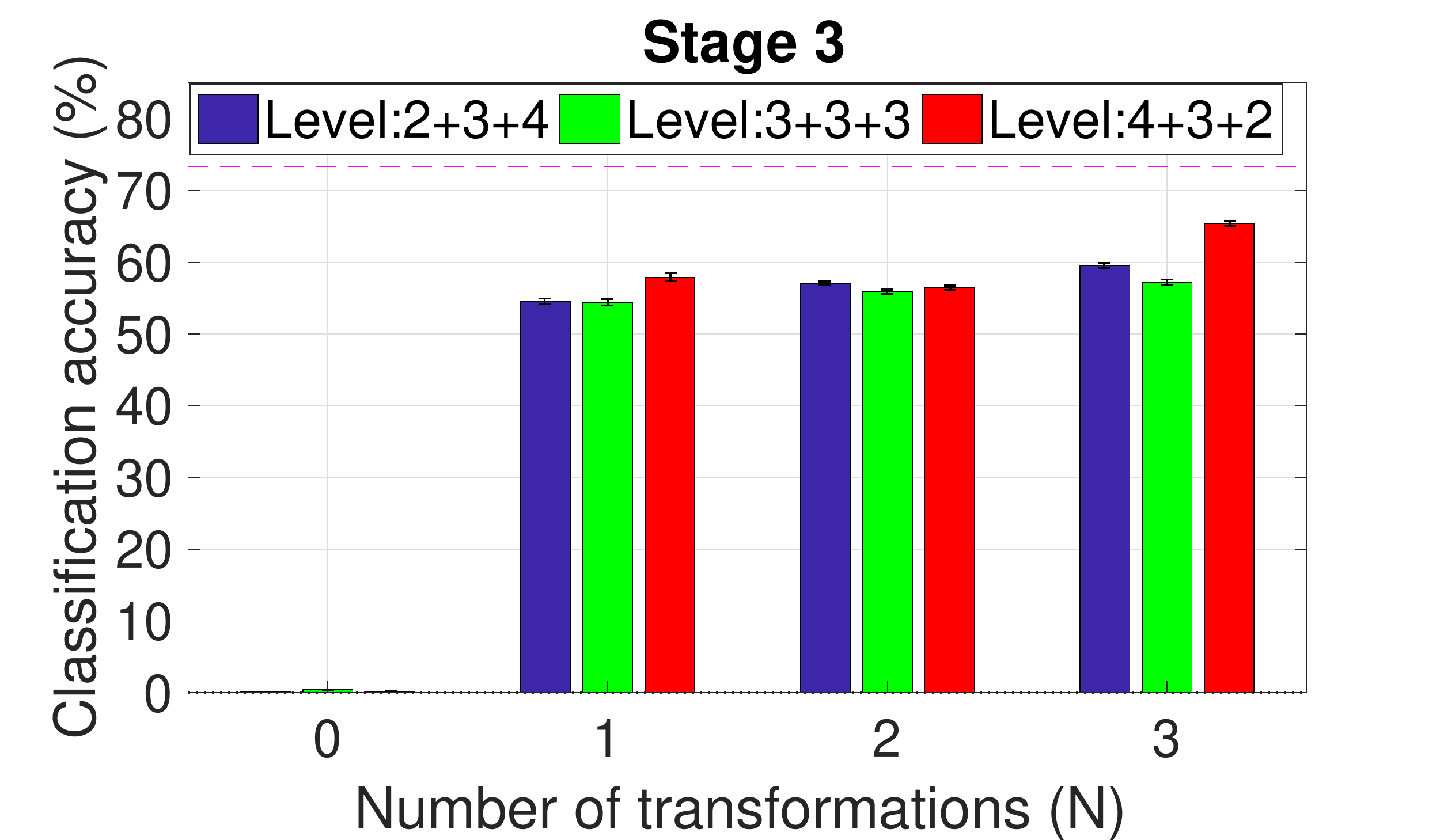}
    }
 \caption{Classification accuracy with reconstructed baseline features from different branching networks on VGG-16 (BN). We run each reconstructed model 3 times and plot the average accuracy and standard deviation.  The magenta dashed line denotes the original baseline accuracy.}
 \label{fig_vgg_complexity}
\end{figure*}

\subsection{Branching Impact on Backbones}
\label{branch_impact}

Using non-linear transformations and knowledge consistency as described in Sec.~\ref{sec:consistency}, we attempt to dig deeper behind the impact of the different branching patterns. We plot in Figs.~\ref{fig_resnet_complexity} and \ref{fig_vgg_complexity} the classification accuracy achieved by the baseline ResNet-18 and VGG-16 when substituting features reconstructed from the different branching networks.  The gap to reach the original baseline accuracy (dashed magenta) is an indirect measurement of non-consistent knowledge that cannot be reconstructed. 

At all three stages, using only a linear transformation ($N\!=\!0$) results in lower accuracies than with non-linear transformations.  This is most prominent at the third stage and is likely due to the difficulty of reconstructing specific semantic abstractions (\eg object parts and shapes) versus low-level features (\eg color and texture) at earlier stages.  Adding more non-linearities usually leads to improvements, but in some cases, a higher $N$ actual results in a lower accuracy (\eg 2+3+4 for Stages 2 and 3). Such an observations is in line with~\cite{liang2019knowledge}, who state that reconstructing shallow layers (\ie earlier stages) was usually more stable than deep layers and that higher $N$ for the deeper layers does not help.

In the first two stages, features from complexity-decreasing branches (4+3+2) seems to be the most consistent with the baseline as they yield the highest classification accuracies.  In the third stage, it is comparable for linear and $N\!=\!1$ while for $N\!=\!2$ and $N\!=\!3$, the other branching patterns do better.  Based on these findings, we conclude that complexity decreasing branches have lower disruptions to the feature abstraction hierarchy of the backbone than the other two branching patterns.  

\section{Conclusion \& Future Work}
In this paper, we have systematically analyzed different branching patterns under fixed total FLOP budgets. We find that using a complexity-decreasing branching pattern achieves the best accuracy-cost trade-off.  Through an indirect estimate of knowledge consistency, we speculate that their effectiveness compared to other branching patterns is due to the preservation of the backbone abstraction hierarchy.  These findings are validated through extensive experimentation on multiple network backbones.  In addition, performance can be further boosted with our weighted ensemble distillation, allowing us to achieve higher accuracy with lower cost compared to current state-of-the-art.  Overall, our complexity-decreasing branching classifiers form a versatile plug-and-play module in multi-exit systems and can improve the performance on various weak or strong and compact or dense backbone architectures.

Although this paper evaluates the effectiveness of complexity-decreasing branch classifiers under the same fixed total FLOP budgets for image classification, more complex pixel-level understanding tasks are unexplored yet. 
In the future, considering more complex structures in the pixel-level tasks (\emph{e.g.}, Autoencoder in semantic segmentation), we will redesign different branching patterns under the fair comparison to investigate their performance and impacts for feature disruption. Therefore, such research can enable the deployment of deep neural networks onto more resource-limited platforms, as well as more generalized AI applications.  



\bibliographystyle{IEEEtran}
\bibliography{reference}

%

\begin{IEEEbiographynophoto}{Shaohui Lin} (Member)
is currently a research professor of East China Normal University, Shanghai, China. 
He received his Ph.D. degree in the School of Informatics from Xiamen University, graduated with Excellent Ph.D Thesis Award in Fujian Province. His research interests include machine learning and computer vision. He has 
severed SPC of IJCAI 2021 and reviewers on several top-tie journals and conference (\eg IEEE TPAMI, TNNLS and CVPR \emph{etc.}) 
\end{IEEEbiographynophoto}

\begin{IEEEbiographynophoto}{Bo Ji}
received the B.E. degree from Zhejiang
University, Hangzhou, China, in 2020. He is
currently pursuing the PhD degree in computer science at National University of Singapore. His current research interests include deep learning and low-level vision.
\end{IEEEbiographynophoto}

\begin{IEEEbiographynophoto}{Rongrong Ji}
(Senior Member, IEEE) is a Nanqiang
Distinguished Professor at Xiamen University, the
Deputy Director of the Office of Science and Technology at Xiamen University, and the Director of Media
Analytics and Computing Lab.
His research falls in the field of computer vision,
multimedia analysis, and machine learning.
He was the
recipient of the Best Paper Award of ACM Multimedia 2011. He has served
as Area Chairs in top-tier conferences such as CVPR and ACM Multimedia.
\end{IEEEbiographynophoto}


\begin{IEEEbiographynophoto}{Angela Yao}
received the BASc degree from the University of Toronto in 2006 and a Ph.D. from ETH Zurich in 2012. She is currently an assistant professor at the School of Computing at the National University of Singapore. Her research interests include computer vision and machine learning and she works on topics in video understanding and human pose estimation.  
\end{IEEEbiographynophoto}




\end{document}